\documentclass[preprint,12pt]{elsarticle}

\usepackage[utf8]{inputenc}
\usepackage[T1]{fontenc}

\usepackage{amsmath}
\usepackage{amssymb}
\usepackage{array}
\usepackage[scaled=0.85]{beramono}
\usepackage{comment}
\usepackage{enumitem}
\usepackage{graphicx}
\usepackage{longtable}
\usepackage{multicol}
\usepackage{multirow}
\usepackage{pdflscape}
\usepackage{xcolor}

\usepackage{caption}

\usepackage{algpseudocode}

\newcounter{algorithm}
\renewcommand{\thealgorithm}{\arabic{algorithm}}
\algrenewcommand\algorithmicindent{1.0em}

\usepackage{listings}
\usepackage{url}
\usepackage{pifont}
\newcommand{\xmark}{\scriptsize\ding{53}\space}
\newcommand{\bcircle}{\scriptsize\ding{108}\space}
\newcommand{\wcircle}{\scriptsize\ding{109}\space}

\journal{Computer Communications}

\begin{document}

\frenchspacing

\begin{frontmatter}

\title{CRAWO: Custom Resources for Adaptive Workload Orchestration}

\author{Eugênio Santos}
\ead{eugenio.lopes.090@ufrn.edu.br}

\author{Daniel Maia}
\ead{daniel.azevedo.095@ufrn.edu.br}

\author{Stefano Loss}
\ead{momoloss10@gmail.com}

\author{José Manoel Silva}
\ead{manoel.freitas.071@ufrn.edu.br}

\author{Aluizio Rocha Neto}
\ead{aluizio@imd.ufrn.br}

\author{Thais Batista}
\ead{thaisbatista@gmail.com}

\author{Everton Cavalcante\corref{corr}}
\ead{everton.cavalcante@ufrn.br}

\author{Nélio Cacho}
\ead{neliocacho@dimap.ufrn.br}

\author{Eduardo Nogueira}
\ead{eduardo@imd.ufrn.br}

\author{Daniel Araújo}
\ead{daniel@imd.ufrn.br}

\author{Frederico Lopes}
\ead{fred@imd.ufrn.br}

\cortext[corr]{Corresponding author}

\affiliation{organization={Federal University of Rio Grande do Norte},
    city={Natal},
    country={Brazil}
}

\begin{abstract}
Edge Intelligence has emerged as a key paradigm for enabling real-time applications in smart cities by shifting computation from centralized cloud data centers to the network edge, thereby reducing latency and bandwidth consumption. However, deploying Artificial Intelligence (AI) pipelines across heterogeneous edge infrastructures remains challenging due to the wide range of device capabilities, from low-power microcontrollers to accelerator-equipped systems. Existing edge orchestration platforms primarily focus on deployment automation and infrastructure management, but these approaches are often inefficient and limit the ability to adaptively allocate resources under dynamic conditions. To tackle these issues, this paper introduces CRAWO (Custom Resources for Adaptive Workload Orchestration), an architectural framework for coordinating AI pipelines across distributed edge environments. CRAWO follows a control-loop-based model that separates allocation intelligence from execution by managing placement decisions, state management, and inter-stage data flows while instantiating services on edge nodes. The framework incorporates a hardware-aware allocator with a pluggable multi-criteria decision layer that leverages real-time infrastructure metrics to enable adaptive workload placement. The reference implementation adopts a microservices architecture deployed on a lightweight Kubernetes distribution (K3s), using Custom Resource Definitions (CRDs) for domain modeling and a dedicated operator for state reconciliation. Evaluation in a vehicle surveillance scenario using license plate recognition demonstrates improved workload distribution and reduced reliance on centralized cloud processing in latency-sensitive environments.
\end{abstract}




\end{frontmatter}


\section{Introduction}
\label{intro}
The shift of computation from centralized cloud data centers to the network edge has positioned edge intelligence as a fundamental paradigm for real-time applications. Edge intelligence enables Artificial Intelligence (AI) processing at or near data sources by leveraging the enhanced computational capabilities of modern edge devices. This approach reduces communication latency, alleviates bandwidth pressure, and enables timely analytics at the source~\cite{singh2023edgeai}, benefits that are often lost in traditional cloud-based approaches. These capabilities are particularly crucial for AI-driven video processing, in which tasks must be performed near the capture devices to meet strict time-sensitive requirements and avoid the prohibitive overhead of transmitting raw, high-volume multimedia streams to a remote infrastructure~\cite{sonnara2025efficient}.

While cloud-only architectures simplify management through a global system view, they are often impractical at scale. For example, in scenarios involving object detection and recognition from images or video streams, even a small number of high-definition cameras can overload the network backhaul, thereby hampering reliable target identification. To address this challenge, complex AI workloads are increasingly modeled as data flows by decomposing monolithic applications into modular pipelines. This structure allows AI services to be expressed as interconnected stages (e.g., detection, recognition, and decision), enabling each component to be deployed on different nodes across the edge-cloud continuum to reduce end-to-end latency and network overhead~\cite{cheng2017fogflow}.

On the other hand, implementing AI pipelines across edge infrastructures remains a significant challenge, as these environments often comprise a wide range of heterogeneous devices, from low-power microcontrollers to units with integrated GPUs and AI accelerators~\cite{singh2023edgeai}. Such diversity makes static deployment strategies inefficient, as each node exhibits distinct constraints and performance characteristics. This challenge is further amplified in scenarios such as smart cities, which demand low latency and efficient coordination across interconnected devices and mobile processing units~\cite{silva2022fogsurvey}. Consequently, an intelligent orchestrator is essential to manage distributed processing. This orchestrator must translate high-level service requirements into concrete execution plans by dynamically allocating computationally intensive workloads encapsulated as microservices or containers to the most suitable nodes. This process must consider resource availability and critical operational constraints to ensure efficiency and scalability under dynamic cluster conditions~\cite{ruiu2025continuum,rochaneto2021thesis,zeydan2022mcdm}.

To address these issues, this work presents CRAWO (\textit{Custom Resources for Adaptive Workload Orchestration}). We designed CRAWO to operate alongside modern workload execution platforms, leveraging their reliability while addressing orchestration challenges in complex edge environments. A key aspect of the solution is the separation of execution and orchestration concerns. While the underlying platform manages service deployment and system state, CRAWO introduces a dedicated layer for allocation and context management. This design enables adaptive coordination without modifying internal execution mechanisms. By treating processing pipelines as cohesive units, CRAWO allows platforms such as Kubernetes to focus on execution consistency while it manages the ``where and when'' of pipeline deployment based on runtime conditions.

The contributions of this work are threefold:
\begin{enumerate}[noitemsep]
    \item A control-loop model organized into different planes, featuring a hardware-aware allocator that employs real-time metrics and a pluggable multi-criteria decision layer for dynamic task distribution.
    \item A reference implementation for CRAWO following a microservices architecture deployed on K3s,\footnote{\url{https://k3s.io}} leveraging Kubernetes Custom Resources Definition (CDR) for domain modeling and a Go-based component for state reconciliation.
    \item A demonstration of our proposal with a license plate recognition (LPR) use case and its performance evaluation considering network conditions and varying high-resolution video workloads.
\end{enumerate}

The remainder of this paper is organized as follows. Section~\ref{useCase} introduces a vehicular surveillance scenario as a motivational use case. Section~\ref{relatedWorks} discusses related work. Section~\ref{crawo} details the CRAWO architecture, while Section~\ref{implementation} describes the reference implementation. Section~\ref{useCaseImplementation} demonstrates the practical instantiation for an LPR use case. Section~\ref{evaluation} presents the evaluation. Section~\ref{conclusion} provides final remarks and directions for future work.

\section{Motivational use case: Vehicular surveillance in smart cities}
\label{useCase}
Our motivational scenario focuses on mobile law enforcement units, specifically police patrol vehicles operating within a smart city environment. These vehicles are equipped with high-definition cameras that continuously record video for monitoring. To transform this raw data into actionable intelligence, the video streams must be processed efficiently to support immediate police decision-making. The critical requirement is to perform LPR in real time, ensuring that suspicious vehicle identification occurs instantly on-site and that video processing does not depend solely on network connectivity.

\begin{figure}[ht]
    \centering 
    \includegraphics[width=0.8\textwidth]{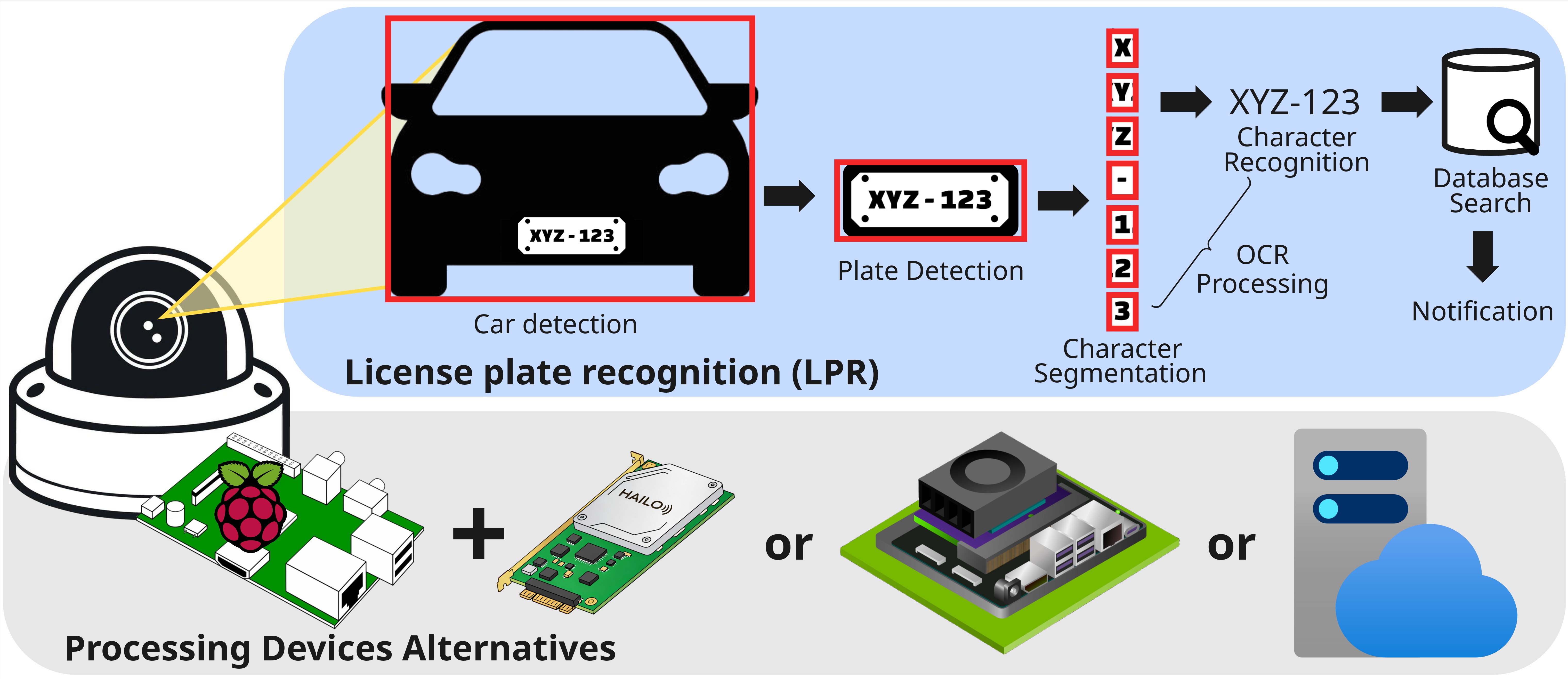} 
    \caption{Use case: Vehicular surveillance in smart cities and processing alternatives.} 
    \label{fig:usecase} 
\end{figure}

As illustrated in \figurename~\ref{fig:usecase}, the LPR workflow consists of a sequential computer vision pipeline. This process begins by detecting the vehicle in raw video frames, followed by localizing and isolating the specific license plate area. Subsequently, the system performs character segmentation to separate individual letters and numbers, followed by Optical Character Recognition (OCR) and database queries to verify the vehicle's legal status.

To execute this pipeline, the vehicular infrastructure is highly heterogeneous, offering various processing options with distinct trade-offs. The need for local processing is driven by the dynamic nature of vehicular surveillance, which demands response times in the order of milliseconds to effectively identify and intercept moving targets. Furthermore, the architecture must be able to seamlessly propagate updated detection rules, such as stolen-vehicle watchlists, from the central cloud to the distributed edge nodes.

\subsection{Cloud-only execution baseline} 
In the conventional cloud-only deployment model, the vehicle serves solely as a data generator, with all stages of the AI pipeline executed remotely on a centralized infrastructure. This approach is commonly adopted in smart city surveillance architectures, as it simplifies management by relying on centralized cloud infrastructure or a single Kubernetes cluster for service execution~\cite{bohm2022cloud}.

However, such centralized processing requires the continuous transmission of high-volume video streams to remote infrastructures, which introduces significant bandwidth pressure and latency. These constraints often violate the real-time requirements of law enforcement applications, as the system remains entirely dependent on the stability and availability of the network backhaul.

\subsection{Edge-distributed execution with orchestration} 
An edge-distributed execution model can offload computationally intensive stages to heterogeneous nodes located within the vehicle, closer to the data source. In this architecture, a microcontroller handles acquisition and virtualization, acting as a device adapter that encapsulates the sensor and standardizes the data flow. Computationally intensive tasks, such as OCR, are then forwarded over the local network to more robust onboard edge nodes, such as an NVIDIA Jetson or a Raspberry Pi.

While this approach significantly reduces latency and backhaul bandwidth consumption, it introduces a trade-off in architectural complexity. Managing a distributed set of heterogeneous mobile nodes is far more demanding than maintaining a centralized cloud because it requires handling intermittent connectivity, hardware-specific constraints, and the overhead of local resource coordination.

To mitigate these challenges, the system's effectiveness becomes entirely dependent on a sophisticated orchestration layer. This orchestrator must meet three critical requirements:

\begin{enumerate}[noitemsep]
    \item \textit{Dynamic workload mapping:} It must automatically assign AI inference tasks to GPU-accelerated nodes while keeping lightweight adaptation tasks on microcontrollers to prevent bottlenecks.
    \item \textit{Resource awareness:} It must maintain real-time awareness of the heterogeneous hardware state to ensure millisecond-range response times and optimize data flows.
    \item \textit{Seamless synchronization:} It must support selective synchronization of metadata to the cloud, ensuring the system remains interoperable with broader smart city ecosystems without overloading the network.
\end{enumerate}

These requirements highlight the limitations of general-purpose orchestration in specialized edge environments. To address these challenges, we propose CRAWO to bridge the gap between high-level AI pipeline definitions and the physical constraints of heterogeneous nodes. By implementing a control-loop model that operates alongside existing execution platforms, CRAWO can transform the complex operational requirements of the LPR scenario into a manageable, automated workflow.

\section{Related work}
\label{relatedWorks}
Orchestration across edge and cloud infrastructures has been widely investigated to support real-time applications. Early efforts, such as FogFlow~\cite{cheng2017fogflow}, introduced programming abstractions for deploying Internet of Things (IoT) service pipelines across distributed resources, enabling application components to be positioned closer to data sources to reduce latency and network overhead. More recent platform-oriented approaches have extended cloud-native orchestration mechanisms to the edge by leveraging Kubernetes as the underlying execution environment. These solutions typically maintain compatibility with the Kubernetes control plane while enabling the deployment of containerized services on geographically distributed, resource-constrained nodes.

Such frameworks provide localized autonomy and facilitate infrastructure management across heterogeneous devices, supporting container execution, remote management, and partial network independence~\cite{bohm2022cloud}. However, the literature on fog and edge computing consistently identifies heterogeneity, mobility, and dynamic resource variability as persistent challenges, particularly in large-scale deployments characterized by fluctuating network conditions and diverse hardware capabilities~\cite{silva2022fogsurvey}. Although existing solutions enable distributed execution, their scheduling logic frequently remains aligned with assumptions inherited from data center environments, where resource stability and network predictability are significantly higher.

\subsection{Workload placement and multi-criteria decision-making}

In many existing systems, workload placement decisions are primarily driven by predefined descriptors such as CPU and memory requests, node labels, or static policies~\cite{bohm2022cloud}. While some solutions incorporate basic network awareness or topology constraints, allocation logic is often tightly coupled to the underlying execution environment and limited to rule-based or single-metric optimization strategies. Under highly dynamic edge conditions, where latency, bandwidth, accelerator availability, and workload intensity continually vary, such approaches often fail to provide sufficiently adaptive decision-making behavior.

To address these limitations, recent research explores intelligent workload placement strategies based on multi-criteria decision-making (MCDM). Methods such as the Technique for Order of Preference by Similarity to Ideal Solution (TOPSIS) and VIKOR have been applied to balance conflicting objectives, including latency, throughput, and resource utilization in distributed resource selection scenarios~\cite{zeydan2022mcdm}. While TOPSIS ranks alternatives based on their geometric distance from an ideal reference solution~\cite{papathanasiou2018topsis}, VIKOR adopts a compromise ranking mechanism that jointly considers aggregate utility and individual regret~\cite{opricovic2004vikor}. This strategy allows for more balanced decisions when criteria exhibit inherent conflicts.

Table~\ref{tab:algorithm_comparison} summarizes representative allocation approaches based on their optimization characteristics and suitability for heterogeneous edge environments. Traditional load-balancing approaches, such as Random, Round Robin, and Least Loaded, either neglect multiple decision criteria or assume homogeneous processing capabilities, thus limiting their effectiveness in heterogeneous edge scenarios. In contrast, MCDM models explicitly capture trade-offs among performance, latency, and resource utilization. Among these approaches, VIKOR stands out for its ability to handle conflicting evaluation criteria in dynamic environments.

\begin{table}[ht]
\centering
\caption{Comparison of representative workload allocation strategies in edge environments.}
\label{tab:algorithm_comparison}
\small
\renewcommand{\arraystretch}{1.15}
\begin{tabular}{>{\raggedright}p{1.5cm}ccc>{\raggedright\arraybackslash}p{3.4cm}}
\hline
Strategy & Multi-criteria & \shortstack{Conflicting\\criteria} & \shortstack{Computational\\complexity} & Edge suitability\\
\hline
Random Selection & \xmark & \xmark & $\mathcal{O}(1)$ & Low (no resource awareness) \\
Round Robin      & \xmark & \xmark & $\mathcal{O}(1)$ & Low (assumes homogeneity) \\
Least Loaded     & \xmark & \xmark & $\mathcal{O}(m)$ & Moderate (single-metric adaptation) \\
TOPSIS           & \bcircle & \wcircle & $\mathcal{O}(mk)$ & High (multi-objective ranking) \\
VIKOR            & \bcircle & \bcircle & $\mathcal{O}(mk)$ & High (compromise-based decision)\\
\hline
\multicolumn{5}{l}{\scriptsize Legend: \bcircle = fully supported; \wcircle = partially supported; \xmark = not supported
}\\[-0.1cm]
\multicolumn{5}{l}{\scriptsize $m$: number of candidate nodes; $k$: number of criteria}\\
\end{tabular}
\end{table}

\subsection{Architectural coupling in edge orchestration}

Recent Kubernetes-based edge orchestration proposals enhance scheduling behavior through tightly integrated plugins and control plane extensions~\cite{rosmaninho2024edgecloud}. Although these approaches improve resource awareness, the allocation logic remains embedded within the underlying environment. Consequently, decision intelligence is implemented at the scheduler level, resulting in strong coupling between execution control and placement strategy. This architectural coupling implies that evolving, replacing, or experimentally evaluating alternative decision models often requires modifications to the orchestration framework itself. Therefore, allocation quality depends not only on deployment capabilities but fundamentally on how decision logic is architecturally positioned. 

Table~\ref{tab:orchestrators_comparison} summarizes various edge-cloud platforms according to the scope of decision criteria supported in their placement logic, as reported in their documentation and the literature~\cite{cheng2017fogflow,bohm2022cloud,silva2022fogsurvey}. We can note that either resource descriptors, network metrics, hardware capability differentiation, or multi-criteria ranking mechanisms are natively incorporated into allocation decisions. While several platforms support container deployment across heterogeneous infrastructures, the combination of explicit multi-criteria ranking with architecturally decoupled allocation intelligence is rarely treated as a primary design principle.

\begin{table}[ht]
\centering
\small
\caption{Decision criteria and allocation strategies adopted by representative edge-cloud orchestrators.}
\label{tab:orchestrators_comparison}
\begin{tabular}{lccccc}
\hline
Platform/framework & CPU & Memory & \shortstack{Network\\metrics} & \shortstack{Hardware\\awareness} & \shortstack{Native\\MCDM} \\
\hline
KubeEdge  & \bcircle & \bcircle & \wcircle & \wcircle & \xmark \\
FogFlow   & \bcircle & \bcircle & \bcircle & \wcircle & \xmark \\
OpenYurt  & \bcircle & \bcircle & \wcircle & \wcircle & \xmark \\
Baetyl    & \bcircle & \bcircle & \wcircle & \wcircle & \xmark \\
CRAWO (this work)
          & \bcircle & \bcircle & \bcircle & \bcircle & \bcircle \\
\hline
\multicolumn{6}{l}{\scriptsize Legend: \bcircle = fully supported; \wcircle = partially supported; \xmark = not supported/detailed}\\
\end{tabular}
\end{table}

In contrast to these embedded approaches, CRAWO introduces an architectural model that explicitly separates allocation intelligence from execution management. The reference implementation adopts a VIKOR-based ranking strategy integrated within a Kubernetes-native control loop. However, the decision mechanism is not bound to a specific algorithm, but rather the allocation layer is designed to support many multi-criteria models without modifying the underlying execution environment. This separation enables systematic experimentation with alternative decision strategies while maintaining architectural consistency across heterogeneous edge deployments.

\section{CRAWO: Custom Resources for Adaptive Workload Orchestration}
\label{crawo}
We designed the CRAWO framework to orchestrate multi-stage AI processing pipelines across heterogeneous edge computing infrastructures. It addresses the challenge of managing complex workloads on distributed nodes with varying capabilities by providing a high-level abstraction layer that decouples service definition from infrastructure complexity.

CRAWO currently works as the orchestrator for the SAALSA (Stream AI Analytics for Live Situational Awareness) framework~\cite{lira2025enhancing,loss2025framework}. While SAALSA provides the underlying intelligent video processing capabilities, CRAWO orchestrates these analytics to achieve high-performance situational awareness in smart city environments.

\subsection{Architecture}
\begin{figure}[ht]
    \centering
    \includegraphics[width=0.7\textwidth]{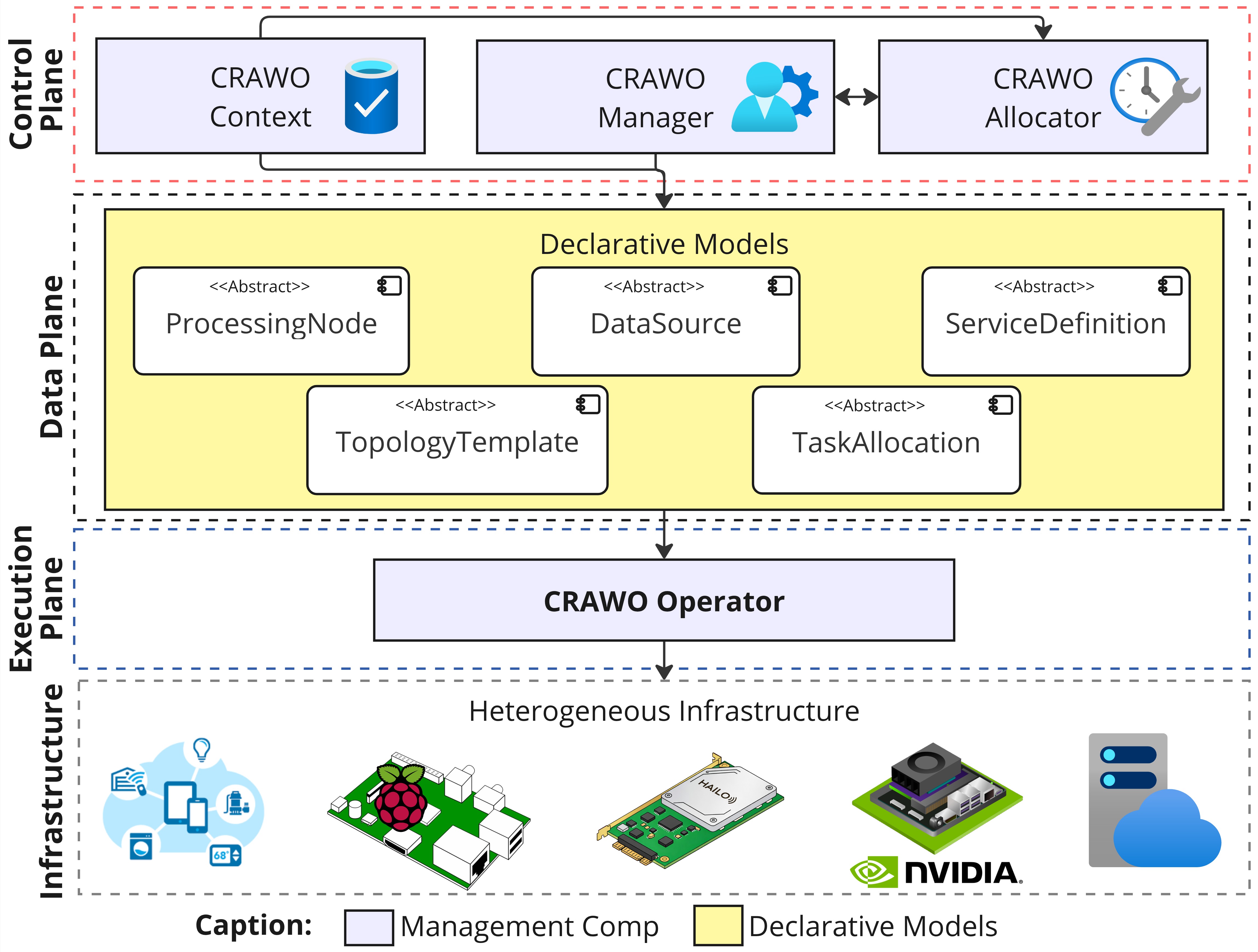}
    \caption{CRAWO architecture.}
    \label{fig:architecture}
\end{figure}

Unlike edge orchestrators that embed decision logic directly within the underlying execution environment, making placement policies inseparable from deployment mechanisms, CRAWO isolates intelligence, domain modeling, and enforcement into three coordinated planes. This separation allows allocation strategies to evolve independently of infrastructure concerns. The CRAWO architecture follows a control-loop model organized into three distinct planes, as depicted in \figurename~\ref{fig:architecture}:

\begin{itemize}[noitemsep]
  \item The \textit{Control Plane} provides the framework’s intelligence, handles user interactions, manages service compositions, and executes decision-making logic to map abstract workloads to physical resources, while maintaining a global view of the domain and optimizing execution strategies based on real-time conditions.
  \item The \textit{Data Plane} defines standardized domain-specific data models for edge nodes, AI services, topologies, and task allocations, serving as the common language that binds the \textit{Control} and \textit{Execution} planes.
  \item The \textit{Execution Plane} continuously monitors the definitions provided by the \textit{Control Plane} and enforces them on the physical infrastructure, managing the life cycle of deployed services to ensure the actual state converges with the desired state.
\end{itemize}

\subsection{Core components}

The architecture is realized through a set of collaborating components, each fulfilling a distinct role in the orchestration lifecycle.

The \textit{Manager} is the orchestration component that coordinates the execution of AI pipelines, maintaining the system's structural aspects, including the association between data sources and processing tasks, the set of available processing nodes, and the definition of service topologies. As the entry point for expressing execution intent, it translates high-level specifications into plans for the underlying platform. By delegating placement decisions to the \textit{Allocator}, the \textit{Manager} focuses on coordination and enforcement.

The \textit{Allocator} determines how execution decisions are derived from current operating conditions. It consolidates the policies and algorithms used to select suitable nodes for each pipeline stage, taking into account hardware capabilities, resource availability, and performance indicators. The \textit{Allocator} operates independently of deployment and execution mechanisms, ensuring that the logic governing allocation decisions is not entangled with infrastructure-level concerns. This separation allows allocation strategies to be adjusted or replaced as needed without affecting the orchestration process coordinated by the \textit{Manager}.

The \textit{Context} component provides a real-time view of the execution environment by aggregating operational data and metrics on node availability and performance to inform allocation decisions. By limiting this component to reflecting runtime behavior rather than participating in decision-making, CRAWO establishes clear boundaries between state observation and orchestration logic.

The \textit{Operator} bridges high-level decisions made by the \textit{Control Plane} and low-level execution on the infrastructure by translating abstract scheduling decisions into concrete resources. Following a reconciliation pattern, the \textit{Operator} continuously monitors the desired state and enforces it by provisioning execution units, configuring environment variables, and establishing network services. It also closes the control loop by observing the lifecycle of running workloads and updating the status of allocation entities.

\subsection{Orchestration and interaction workflow}

The orchestration process is implemented through a synchronized interaction sequence among the control and execution components. This workflow translates high-level user intents into hardware-aware execution units while maintaining continuous state reconciliation.

\begin{figure}[ht]
    \centering
    \includegraphics[width=\textwidth]{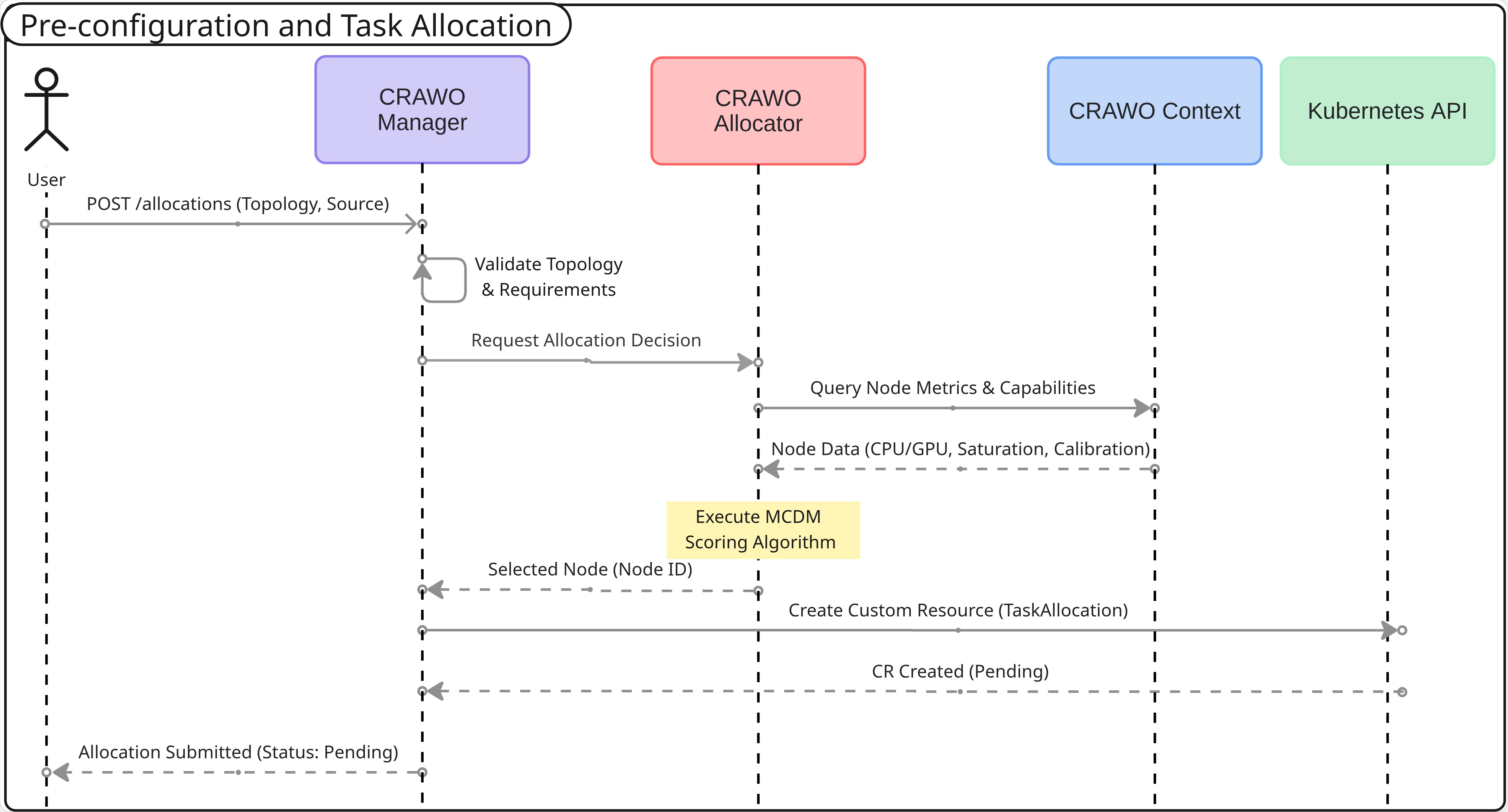}
    \caption{Sequence diagram illustrating the allocation decision flow, from user request to creating a \texttt{TaskAllocation} CR.}
    \label{fig:seq_allocation}
\end{figure}

\textbf{Allocation decision flow.}
The workflow begins when a service request is submitted to the \textit{Manager}. As illustrated in \figurename~\ref{fig:seq_allocation}, the \textit{Manager} validates the topology and requirements before delegating the placement decision to the \textit{Allocator}. The \textit{Allocator} performs a hardware-aware selection by querying the \textit{Context} component for real-time metrics and capabilities. Using a VIKOR-based multi-criteria strategy, it ranks candidate nodes and selects a compromise execution target. Finally, the \textit{Manager} persists this intent as a \texttt{TaskAllocation} Custom Resource (CR) in the Kubernetes API.

\begin{figure}[ht]
    \centering
    \includegraphics[width=0.75\textwidth]{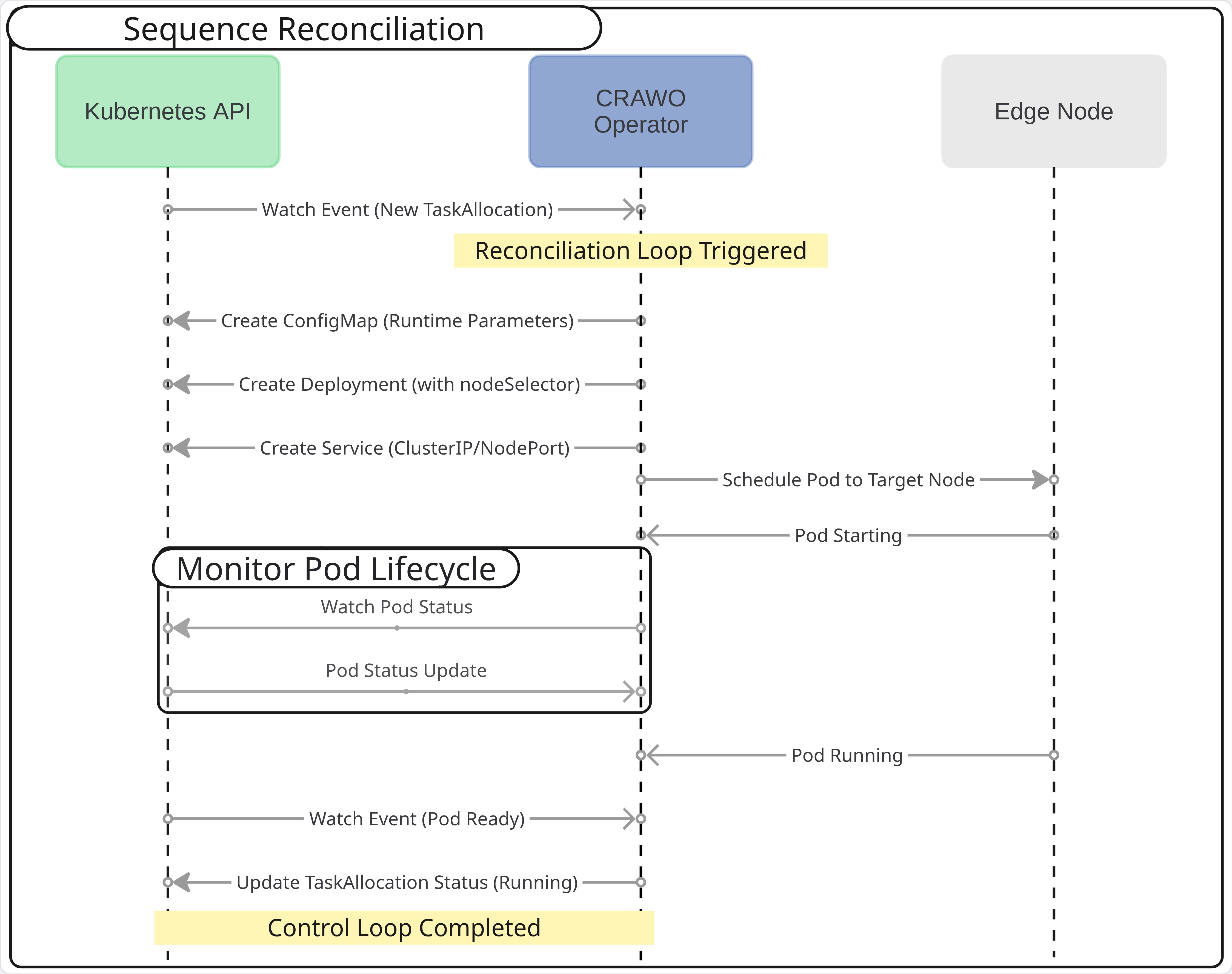}
    \caption{Sequence diagram illustrating the reconciliation flow, from \texttt{TaskAllocation} detection to status synchronization.}
    \label{fig:seq_reconciliation}
\end{figure}

\textbf{Reconciliation and provisioning flow.}
Once the \texttt{TaskAllocation} CR is persisted, the \textit{Execution Plane} takes control. As shown in \figurename~\ref{fig:seq_reconciliation}, the \textit{Operator} detects the new resource via a watch mechanism and initiates the reconciliation loop. It provisions the necessary Kubernetes resources, including a \textit{ConfigMap} for runtime parameters, a \textit{Deployment} for the service container, and, optionally, a \textit{Service} for network access. Crucially, the Operator injects a \texttt{nodeSelector} constraint to pin the workload to the node selected by the \textit{Allocator}. Throughout the Pod lifecycle, the \textit{Operator} continuously monitors the workload state and reflects it back into the \texttt{TaskAllocation} status field.

\begin{figure}[ht]
    \centering
    \includegraphics[width=0.95\textwidth]{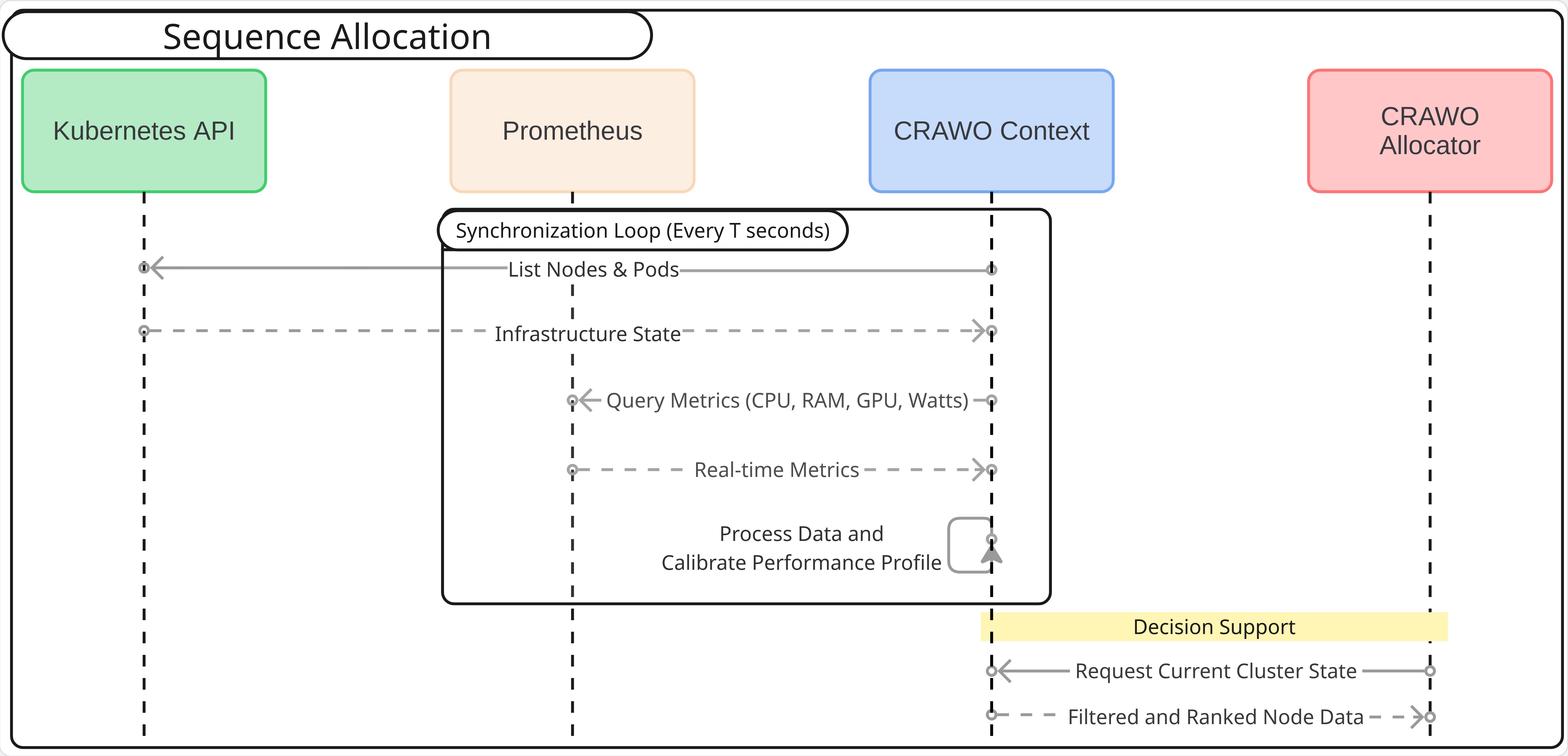}
    \caption{Sequence diagram illustrating the context awareness flow, and how real-time metrics influence allocation logic.}
    \label{fig:seq_context}
\end{figure}

\textbf{Context awareness and resource monitoring.}
A continuous secondary workflow ensures the orchestration engine remains aware of cluster health and performance. The \textit{Context} component periodically pulls metrics from Prometheus (e.g., CPU, RAM, and GPU utilization) and state information from the Kubernetes API. As shown in \figurename~\ref{fig:seq_context}, these metrics are processed and provided to the \textit{Allocator} for calculating scores for future scheduling decisions, ensuring the system adapts to the dynamic conditions of the edge environment.

\subsection{Allocation decision model}
CRAWO introduces an explicit decision layer that assigns AI pipeline stages to heterogeneous edge processing nodes. This layer is conceptually separated from orchestration control and execution concerns, allowing placement decisions to be derived from runtime conditions without coupling them to deployment mechanisms. Within this model, the \textit{Allocator} evaluates candidate nodes and selects an execution target for each pipeline stage based on a structured sequence.

First, the \textit{Allocator} identifies the set of processing nodes that satisfy basic feasibility constraints, such as architectural compatibility, reachability, and execution capacity. This filtering step ensures that only nodes capable of executing the service are considered. From this reduced candidate set, the \textit{Allocator} builds a decision matrix using runtime information provided by the \textit{Context} component.

\subsubsection{Decision criteria and selection strategy}
The decision criteria reflect both performance and resource-utilization aspects relevant to heterogeneous edge environments. In the current implementation, these criteria include:

\begin{itemize}[noitemsep]
    \item Inference throughput (frames per second, FPS): it is prioritized as a benefit criterion to sustain real-time processing rates and avoid backlog.
    \item Estimated processing latency: it is treated as a cost criterion to prevent violations of strict response-time constraints.
    \item Node load: it is treated as a cost criterion to prevent resource saturation and ensure stability under multi-tenant conditions.
\end{itemize}

Together, these indicators capture the inherent trade-off between maximizing performance and avoiding resource exhaustion. To rank candidate nodes according to these criteria, the reference implementation for CRAWO uses the VIKOR method~\cite{opricovic2004vikor}. This MCDM technique identifies a compromise solution by simultaneously minimizing group utility loss and individual regret. Its compromise-ranking formulation enables balanced decisions without privileging any single performance dimension, making it more robust than common baselines. For instance, Random selection ignores node conditions and may overload constrained devices, Round Robin assumes homogeneous capabilities, and Least Loaded disregards differences in hardware acceleration. By jointly accounting for throughput, latency, and node load, the VIKOR method yields a more robust compromise solution under dynamic and heterogeneous edge conditions.

The VIKOR-based allocation strategy evaluates each candidate node against the three runtime decision criteria that capture the fundamental trade-off between computational performance and resource saturation in heterogeneous edge environments. Within the VIKOR method, these criteria jointly contribute to the computation of a compromise ranking that balances overall utility and individual regret. Table~\ref{tab:criteria} summarizes the adopted criteria, their optimization direction, and the relative weights used in the current implementation of the \textit{Allocator}.

\begin{table}[ht]
\centering
\footnotesize
\caption{Decision criteria adopted by the \textit{Allocator} in CRAWO.}
\label{tab:criteria}
\begin{tabular}{llcl}
\hline
Criterion & Type & Weight & Interpretation \\
\hline
FPS & Benefit (max) & 0.40 & Inference throughput (processing performance) \\
Latency & Cost (min) & 0.35 & Estimated network delay to execute the stage \\
Load & Cost (min) & 0.25 & Current node utilization level (congestion) \\
\hline
\end{tabular}
\end{table}

\subsubsection{Mathematical formulation}
The VIKOR method determines a compromise solution by evaluating the relative distance of each alternative from the ideal performance for each criterion. Initially, a decision matrix $F = [f_{ij}]$ is defined, where each candidate node $i$ is evaluated against criterion $j$. For each criterion, the best value $f_j^{*}$ and the worst value $f_j^{-}$ are identified. The limits are defined by Equation~\ref{eq:benefit} for benefit criteria (to be maximized) and follow Equation~\ref{eq:cost} for cost criteria (to be minimized).

\begin{align}
f_j^{*} &= \max_i f_{ij}, \quad f_j^{-} = \min_i f_{ij} \label{eq:benefit} \\
f_j^{*} &= \min_i f_{ij}, \quad f_j^{-} = \max_i f_{ij} \label{eq:cost}
\end{align}

Once the ideal points are established, the normalized distance $d_{ij}$ of an alternative $i$ under criterion $j$ is computed. For benefit and cost criteria, these distances are defined as shown in Equation~\ref{eq:dist_cost}. These values are subsequently aggregated using predefined weights $w_j$ to derive the group utility $S_i$ and the maximum individual regret $R_i$, according to the formulations in Equation~\ref{eq:S_R_measures}:

\begin{equation}
\label{eq:dist_cost}
\text{Benefit: } d_{ij} = \frac{f_j^{*} - f_{ij}}{f_j^{*} - f_j^{-}} \quad 
\text{Cost: } d_{ij} = \frac{f_{ij} - f_j^{*}}{f_j^{-} - f_j^{*}}
\end{equation}

\begin{equation}
\label{eq:S_R_measures}S_i = \sum_{j=1}^{n} w_j d_{ij}, \quad R_i = \max_{j} (w_j d_{ij})
\end{equation}

We defined the adopted weights empirically based on workload requirements observed during controlled experiments in the LPR use case. In the context of LPR, maximizing throughput (0.40) is prioritized to prevent frame dropping at the source, while latency (0.35) remains critical for timely police response, and node load (0.25) serves as a stabilizing penalty rather than a primary driver. The configuration prioritizes sustained inference throughput while maintaining latency constraints and preventing resource saturation. The weight distribution is not fixed by the architectural design and can be adjusted to accommodate different application domains or operational priorities without requiring modifications to the orchestration framework.

The final selection is based on the compromise ranking index $Q_i$, which balances the overall group utility and individual regret. Let $S^{*}, S^{-}, R^{*},$ and $R^{-}$ be the minimum and maximum values of $S_i$ and $R_i$, respectively, as shown in Equation~\ref{eq:extremos}. The index $Q_i$ is then formulated using a strategy weight $v \in [0,1]$ (Equation~\ref{eq:Qi}), where $v = 0.5$ is typically adopted to provide a neutral balance. In cases where the denominators $(S^{-} - S^{*})$ or $(R^{-} - R^{*})$ are zero, the corresponding term is set to zero to prevent division-by-zero errors.

\begin{equation}\label{eq:extremos}
S^{*} = \min_i S_i, \quad S^{-} = \max_i S_i, \quad R^{*} = \min_i R_i, \quad R^{-} = \max_i R_i
\end{equation}

\begin{equation}\label{eq:Qi}
Q_i = v \frac{S_i - S^{*}}{S^{-} - S^{*}} + (1 - v) \frac{R_i - R^{*}}{R^{-} - R^{*}}
\end{equation}

Alternatives are evaluated based on their $Q_i$ values, and the node with the smallest $Q_i$ is selected as the compromise solution. Algorithm~\ref{alg:vikor_allocator} details this mechanism encapsulated within the \textit{Allocator}, thus preserving the policy-agnostic nature of CRAWO. This separation enables the current VIKOR-based strategy to be replaced by alternative heuristics or learning-based approaches without modifications to the orchestration control flow.

\refstepcounter{algorithm}
\begin{small}
\begin{center}
\noindent Algorithm \thealgorithm: VIKOR-based hardware-aware node selection
\label{alg:vikor_allocator}
\end{center}

\begin{algorithmic}[1]
\Require Candidate set $N \neq \varnothing$
\Ensure Selected node $n_{best}$

\State $m \gets |N|$ \hspace*{0.5cm} \Comment{number of candidates}
\State $k \gets 3$ \hspace*{1cm} \Comment{criteria: FPS, latency, load}
\State Initialize matrix $F[m][k]$

\Statex
\Statex{\Comment{\textit{Step 1: Construct decision matrix $F$}}}
\For{$i = 1$ to $m$}
    \State $F[i,1] \gets \mathrm{FPS}(n_i)$
    \State $F[i,2] \gets \mathrm{Latency}(n_i)$
    \State $F[i,3] \gets \mathrm{Load}(n_i)$
\EndFor

\Statex
\Statex{\Comment{\textit{Step 2: Determine best and worst values $f^*$ and $f^-$}}}
\For{$j = 1$ to $k$}
    \If{criterion $j$ is benefit}
        \State $f^*[j] \gets \max_i F[i,j]$
        \State $f^-[j] \gets \min_i F[i,j]$
    \Else
        \State $f^*[j] \gets \min_i F[i,j]$
        \State $f^-[j] \gets \max_i F[i,j]$
    \EndIf
\EndFor

\Statex
\Statex{\Comment{\textit{Step 3: Compute normalized distances $d$}}}
\For{$i = 1$ to $m$}
    \For{$j = 1$ to $k$}
        \If{$f^*[j] = f^-[j]$}
            \State $d[i,j] \gets 0$
        \ElsIf{criterion $j$ is benefit}
            \State $d[i,j] \gets 
            \dfrac{f^*[j] - F[i,j]}{f^*[j] - f^-[j]}$
        \Else
            \State $d[i,j] \gets 
            \dfrac{F[i,j] - f^*[j]}{f^-[j] - f^*[j]}$
        \EndIf
    \EndFor
\EndFor

\Statex
\Statex{\Comment{\textit{Step 4: Compute $S_i$ and $R_i$}}}
\State $W \gets [0.40, 0.35, 0.25]$
\For{$i = 1$ to $m$}
    \State $S[i] \gets \sum_j W[j] \cdot d[i,j]$
    \State $R[i] \gets \max_j W[j] \cdot d[i,j]$
\EndFor

\Statex
\Statex{\Comment{\textit{Step 5: Compute compromise index $Q_i$}}}
\State $S^* \gets \min_i S[i]$
\State $S^- \gets \max_i S[i]$
\State $R^* \gets \min_i R[i]$
\State $R^- \gets \max_i R[i]$
\State $v \gets 0.5$

\For{$i = 1$ to $m$}
    \State $termS \gets 
    \begin{cases}
    \dfrac{S[i] - S^*}{S^- - S^*}, & S^- \neq S^* \\
    0, & \text{otherwise}
    \end{cases}$

    \State $termR \gets 
    \begin{cases}
    \dfrac{R[i] - R^*}{R^- - R^*}, & R^- \neq R^* \\
    0, & \text{otherwise}
    \end{cases}$

    \State $Q[i] \gets v \cdot termS + (1 - v) \cdot termR$
\EndFor

\Statex
\Return $n_{best} \gets \arg\min_i Q[i]$
\vspace*{0.5cm}
\end{algorithmic}
\end{small}

The computational complexity of the VIKOR scoring procedure is $\mathcal{O}(mk)$ for computing the $S_i$ and $R_i$ measures, where $m$ is the number of candidate nodes and $k$ is the number of criteria. The subsequent computation of the compromise index $Q_i$ and the selection of the minimum value require $\mathcal{O}(m)$ time, resulting in an overall linear complexity. In typical edge-cluster scenarios, where the number of candidate nodes per allocation decision is limited, this cost remains negligible.

\section{CRAWO implementation}
\label{implementation}
We developed the reference implementation of the CRAWO platform as a distributed system following the microservices architecture pattern. To ensure portability and scalability across edge environments, the components are containerized and deployed on K3s, a lightweight Kubernetes distribution certified by the Cloud Native Computing Foundation (CNCF) and specifically designed for resource-constrained clusters. K3s preserves full compatibility with the Kubernetes API while reducing the system footprint by consolidating components and minimizing runtime dependencies, making it well-suited for edge deployments.

\subsection{Control Plane}

The \textit{Control Plane} comprises three core microservices implemented in Java 17 using Spring Boot 3.0. This stack was selected for its mature ecosystem and strong type safety, which are essential for managing complex domain models in distributed systems.

The \textit{Manager} acts as the entry point and API Gateway. In the current implementation, it utilizes a PostgreSQL database to persist infrastructure descriptions and pipeline topologies. While the CRAWO architecture conceptually models all entities as native Kubernetes CRDs, a relational database is used to accelerate validation of management logic. The textit{Manager} translates high-level user requests into strongly typed \texttt{TaskAllocation}CRs, serving as the primary interface between the \textit{Control Plane} and the \textit{Execution Plane}.

The \textit{Allocator} is logically decoupled from the management layer and encapsulates the framework’s core decision-making logic. It exposes a RESTful interface to process allocation requests using the VIKOR MCDM method. The decision process balances performance (throughput) against resource constraints (latency and node load). By relying on data from the \textit{Context} component, the \textit{Allocator} remains independent of direct interactions with the infrastructure.

The \textit{Context} component serves as the framework’s information repository. It aggregates static hardware descriptions, calibrated performance profiles, and real-time operational metrics (e.g., node availability and resource utilization). This component provides the situational awareness required for the \textit{Allocator} to perform hardware-aware, adaptive scheduling.

\subsection{Data Plane}
The \textit{Data Plane} state is modeled using Kubernetes CRDs that extend the Kubernetes API to support CRAWO’s domain-specific requirements. These CRDs establish a formal contract for state synchronization between the \textit{Control Plane} and the \textit{Execution Plane}.

The infrastructure and inputs are represented by the \texttt{ProcessingNode} and \texttt{DataSource} resources. \texttt{ProcessingNode} encapsulates the state of edge compute units (e.g., NVIDIA Jetson, Raspberry Pi), storing hardware capabilities (e.g., CPU, RAM, and GPU availability) and scheduling labels. \texttt{DataSource} models input streams, such as RTSP camera feeds, and encapsulates the connection URIs and authentication credentials required for data consumption.

The software catalog is defined through \texttt{ServiceDefinition} and \texttt{Topology\-Template} resources. The former defines individual pipeline components, specifying container images and environment variables, while the latter arranges these services into a directed acyclic graph to represent complex workflows.

Finally, the \texttt{TaskAllocation} serves as the central artifact of the orchestration process. It binds a \texttt{TopologyTemplate} to a specific \texttt{ProcessingNode} and \texttt{DataSource}, thereby declaring the desired state of a running pipeline and triggering the reconciliation loop.

\subsection{Execution Plane}
The execution logic resides in the \textit{Operator}, a custom Kubernetes controller developed in Go using the Kubebuilder framework.\footnote{\url{https://github.com/kubernetes-sigs/kubebuilder}} This choice leverages Go's native efficiency and Kubebuilder’s code-generation capabilities for robust Kubernetes API interactions.

The Operator employs a watch mechanism on \texttt{TaskAllocation} resources to trigger an idempotent reconciliation sequence. For each pipeline stage, the controller:

\begin{enumerate}[noitemsep]
    \item Generates a \textit{ConfigMap} containing runtime parameters
    \item Instantiates a \textit{Deployment} using the specified container image.
    \item Injects a \texttt{nodeSelector} to pin the workload to the specific edge node selected by the \textit{Allocator}
    \item Provisions a \textit{Service} (ClusterIP, NodePort, or LoadBalancer) to ensure network accessibility
\end{enumerate}

The control loop is closed by continuously monitoring the created resources and reflecting their aggregated state (e.g., pending, running, or failed) back into the \texttt{TaskAllocation} status field, providing real-time feedback to the \textit{Control Plane}.

\section{LPR use case using CRAWO}
\label{useCaseImplementation}
This section demonstrates the practical application of the CRAWO framework by instantiating the vehicular surveillance use case introduced in Section~\ref{useCase}. This LPR solution is being integrated into CAD-Patrol~\cite{loss2025sistema}, a mobile system designed to optimize public safety incident response and enhance situational awareness for officers in the field. By leveraging 5G connectivity, edge computing, and AI, CAD-Patrol enables real-time analysis of video captured by smartphone-based cameras. Processing is performed directly on the vehicle’s onboard unit at the network edge, ensuring low latency and operational continuity even in environments with unstable backhaul. 

CAD-Patrol is developed using the SAALSA framework~\cite{lira2025enhancing,loss2025framework}, which provides the underlying intelligent video processing, and utilizes CRAWO as its primary orchestrator. Both solutions are part of the SPICI,\footnote{\url{https://smlab.imd.ufrn.br/spici/en}} an R\&D project aimed at developing integrated public safety applications leveraging 5G networks. These applications have strict requirements for reliable communication, high availability, and low latency in information transfer. Consequently, they require a 5G network infrastructure that provides high capacity and communication speed, supporting a massive number of devices with minimal latency.

To instantiate CRAWO, the LPR pipeline stages are modeled as a set of CRD manifests. Each resource addresses a specific orchestration concern:

\begin{itemize}[noitemsep]
    \item \texttt{ServiceDefinition}: describes a containerized pipeline stage.
    \item \texttt{TopologyTemplate}: encodes the processing directed acyclic graph and its hardware requirements.
    \item \texttt{ProcessingNode}: registers the physical edge infrastructure.
    \item \texttt{DataSource}: binds the video feed (e.g., RTSP) to the pipeline entry point.
    \item \texttt{TaskAllocation}: records the placement decision produced by the \textit{Allocator}.
\end{itemize}

\subsection{Pipeline definition and implementation}
\begin{figure}[ht]
    \centering
    \includegraphics[width=0.9\textwidth]{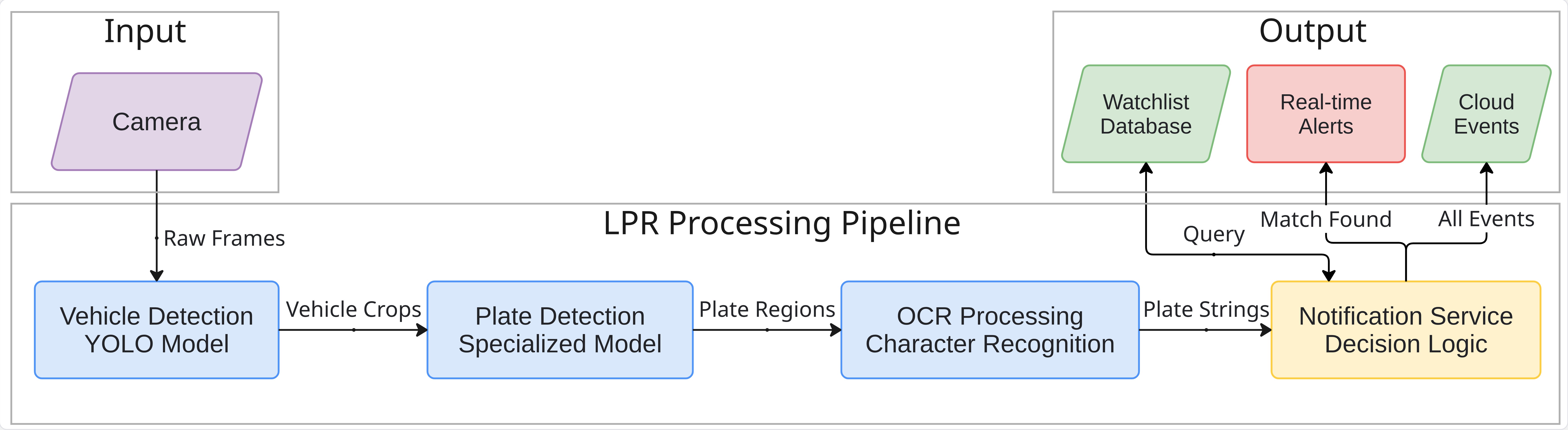}
    \caption{Four-stage LPR pipeline}
    \label{fig:lpr_pipeline}
\end{figure}

The LPR pipeline is decomposed into four sequential stages, following a cascaded detection-recognition pattern: Vehicle Detection, Plate Detection, OCR Processing, and Notification (database verification). The first three stages require GPU acceleration for real-time inference, while the notification stage performs lightweight lookups that can be handled by any CPU-based node. \figurename~\ref{fig:lpr_pipeline}. shows the pipeline dataflow structure.

The \texttt{TopologyTemplate} (see Listing~\ref{lst:topology}) encodes these relationships. The \texttt{hardwareConstraints} block at each stage is the only placement directive required from the human operator. The \textit{Allocator} uses these constraints, along with real-time node metrics, to determine optimal assignments, removing the need for manual node selection.

\begin{lstlisting}[caption={\texttt{TopologyTemplate} for the LPR pipeline}, label={lst:topology}, basicstyle=\footnotesize\ttfamily]
    apiVersion: orchestration.saalsa.spici/v1
    kind: TopologyTemplate
    metadata:
      name: lpr-pipeline
    spec:
      description: "License Plate Recognition Pipeline"
      stages:
        - name: vehicle-detection
          serviceRef: vehicle-detector
          hardwareConstraints:
            gpu: true
            minMemoryMB: 2048
        - name: plate-detection
          serviceRef: plate-detector
          hardwareConstraints:
            gpu: true
            minMemoryMB: 1024
        - name: ocr-processing
          serviceRef: plate-ocr
          hardwareConstraints:
            gpu: true
            minMemoryMB: 1024
        - name: notification
          serviceRef: notification-service
          hardwareConstraints:
            gpu: false
            minMemoryMB: 512
\end{lstlisting}

The video source is declared as a separate \texttt{DataSource} resource (see Listing~\ref{lst:datasource}), which binds the RTSP stream from the on-board camera to the pipeline entry point. Decoupling the data source from the processing graph allows the same \texttt{TopologyTemplate} to be reused across different patrol vehicles without modifying the core pipeline definition. 

\begin{lstlisting}[caption={\texttt{DataSource} binding the patrol vehicle's camera to the LPR pipeline}, label={lst:datasource}, basicstyle=\footnotesize\ttfamily]
    apiVersion: orchestration.saalsa.spici/v1
    kind: DataSource
    metadata:
      name: camera-patrol-01
    spec:
      protocol: rtsp
      uri: "rtsp://192.168.1.10:554/stream"
      pipelineRef: lpr-pipeline
      entryStage: vehicle-detection
\end{lstlisting}

Finally, the heterogeneous nodes aboard the vehicle are registered as \texttt{ProcessingNoderesources} (see Listing~\ref{lst:nodes}), providing the \textit{Allocator} with a stable hardware profile (GPU availability, core count, memory) for constraint filtering and eliminating runtime capability queries during scheduling.

\begin{lstlisting}[caption={\texttt{ProcessingNode} definitions for the vehicular cluster}, label={lst:nodes}, basicstyle=\footnotesize\ttfamily]
    apiVersion: orchestration.saalsa.spici/v1
    kind: ProcessingNode
    metadata:
      name: patrol-01-rpi
    spec:
      hostname: "rpi-patrol-01"
      capabilities:
        cpu: "ARM Cortex-A72"
        cores: 4
        memoryMB: 4096
        gpu: false
      labels:
        vehicle: "patrol-01"
        role: "acquisition"
    ---
    apiVersion: orchestration.saalsa.spici/v1
    kind: ProcessingNode
    metadata:
      name: patrol-01-jetson
    spec:
      hostname: "jetson-patrol-01"
      capabilities:
        cpu: "ARM Cortex-A78AE"
        cores: 12
        memoryMB: 32768
        gpu: true
        gpuModel: "NVIDIA Ampere"
        cudaCores: 2048
      labels:
        vehicle: "patrol-01"
        role: "inference"
\end{lstlisting}

\subsection{Allocation decision and orchestration}
Upon receiving a deployment request, the \textit{Manager} validates the \texttt{Topology\-Template} against the available \texttt{ProcessingNode} pool and invokes the \textit{Allocator}. For the three GPU-dependent stages, the hardware filter immediately identifies \texttt{patrol-01-jetson}as the only viable candidate, so no multi-criteria ranking is required.

For the Notification stage, both the NVIDIA Jetson and the Raspberry Pi devices (\texttt{patrol-01-rpi}) meet the basic CPU and memory constraints. In this case, the VIKOR-based mechanism constructs a decision matrix evaluating processing capacity (IPT, benefit weight 0.40),  network Latency (cost weight 0.35), and current processor load (cost weight 0.25). If the NVIDIA Jetson is heavily committed to the preceding inference stages, its load score is penalized, allowing the lightly loaded Raspberry Pi to rank higher for this CPU-bound task. The resulting placement is persisted as a \texttt{TaskAllocation} resource (see Listing~\ref{lst:taskalloc}), serving as the execution contract between the \textit{Control Plane} and the \textit{Execution Plane}.

\begin{lstlisting}[caption={\texttt{TaskAllocation} produced by the \textit{Allocator} for the LPR pipeline}, label={lst:taskalloc}, basicstyle=\footnotesize\ttfamily]
    apiVersion: orchestration.saalsa.spici/v1
    kind: TaskAllocation
    metadata:
      name: lpr-pipeline-alloc-v1
    spec:
      pipelineRef: lpr-pipeline
      assignments:
        - stage: vehicle-detection
          nodeSelector:
            kubernetes.io/hostname: jetson-patrol-01
        - stage: plate-detection
          nodeSelector:
            kubernetes.io/hostname: jetson-patrol-01
        - stage: ocr-processing
          nodeSelector:
            kubernetes.io/hostname: jetson-patrol-01
        - stage: notification
          nodeSelector:
            kubernetes.io/hostname: rpi-patrol-01
\end{lstlisting}

Upon persistence of the \texttt{TaskAllocation}, the \textit{Operator} then initiates the reconciliation loop described in Section~\ref{implementation}. For each stage, it provisions a \textit{ConfigMap} for runtime parameters, a \textit{Deployment} with the designated \texttt{nodeSelector}, and a \textit{Service} for inter-stage communication. This translates the declarative allocation into a running inference pipeline with no further operator intervention.

\section{Evaluation}
\label{evaluation}
This section presents a systematic evaluation of CRAWO through discrete-event simulation using YAFS (Yet Another Fog Simulator)~\cite{lera2019yafs}, a Python-based framework for fog computing environments. To evaluate the performance and viability of the framework, we formulate three research questions that address the fundamental trade-offs in edge orchestration:

\begin{enumerate}[label={RQ\arabic* --},leftmargin=*]
    \item Decision quality and efficacy: How does the VIKOR-based allocation strategy compare to traditional heuristics and the TOPSIS algorithm regarding end-to-end latency, load balancing, and resource utilization? Furthermore, how sensitive is this performance advantage to increases in workload intensity?
    \item Scalability and network sensitivity: To what extent does the framework maintain stable performance as the infrastructure scales, and how do varying 5G network conditions impact its coordination efficiency?
    \item Deployment architecture: What are the quantifiable benefits of the CRAWO edge-distributed execution model in terms of latency reduction and bandwidth preservation when compared to a conventional cloud-only baseline? 
\end{enumerate}

We detail the experimental process in the following. Section~\ref{sec:eval_metrics} defines performance metrics. Section~\ref{sec:sim_env} describes the simulation environment in terms of topology, application, and network models. Section~\ref{sec:exp_design} presents the experimental design. Section~\ref{sec:sim_impl} details the simulation implementation. Results appear in Section~\ref{sec:results}, followed by a discussion in Section~\ref{sec:discussion}.

\subsection{Evaluation metrics}
\label{sec:eval_metrics}
Four complementary metrics are employed to characterize the allocation strategies from distinct performance perspectives, as summarized in Table~\ref{tab:eval_metrics}:

\begin{itemize}[noitemsep]
  \item \textit{End-to-end latency} ($L_{e2e}$): it measures total elapsed time from frame capture at the source to notification delivery at the sink, capturing user-perceived responsiveness.
  \item \textit{Processing time} ($T_{proc}$): it isolates computational efficiency by aggregating execution time across all pipeline stages, excluding network overhead.
  \item \textit{Network delay} ($T_{net}$): it quantifies infrastructure impact as the cumulative transmission and queuing time across all inter-node hops.
  \item \textit{Resource utilization} ($U$): it measures the effectiveness of workload distribution as the mean fraction of computational capacity actively engaged across all edge nodes.
\end{itemize}

For each metric, we report the mean and 95\% confidence interval across ten independent simulation runs with different random seeds to ensure statistical significance.

\begin{table}[ht]
\centering
\small
\caption{Performance metrics}
\label{tab:eval_metrics}
\renewcommand{\arraystretch}{1.15}
\begin{tabular}{>{\raggedright}p{2cm}l>{\raggedright\arraybackslash}p{9.2cm}}
\hline
Metric & Symbol & Definition \\
\hline
End-to-end latency   & $L_{e2e}$  & Total elapsed time from frame capture at source to notification delivery at sink (ms) \\
Processing time      & $T_{proc}$ & Cumulative execution time across all pipeline stages, excluding network contributions (ms) \\
Network delay        & $T_{net}$  & Cumulative transmission and queuing time across all inter-node hops (ms) \\
Resource utilization & $U$        & Mean fraction of available computational capacity actively engaged across edge nodes (\%) \\
\hline
\end{tabular}
\end{table}

\subsection{Simulation environment}
\label{sec:sim_env}
We utilize YAFS v3.1 to model computational resources, network links, and application workloads as discrete events, enabling precise measurement of latency distributions and resource utilization patterns under controlled conditions. The adoption of a simulation environment was a deliberate methodological choice aimed at isolating the pure decision-making behavior of the allocation algorithms from hardware-induced noise. Real-world edge deployments are subject to confounding factors not captured by the model, including operating-system scheduling jitter, thermal throttling of edge accelerators under sustained load, flash-storage I/O bottlenecks, and non-deterministic 5G handoff events. By parameterizing the infrastructure as a discrete-event system with deterministic resource attributes, YAFS establishes a clear theoretical performance bound that allows the correctness and relative efficacy of the VIKOR-based decision logic to be rigorously validated. This controlled baseline serves as an essential prerequisite for the physical deployment validation planned as future work (see Section~\protect\ref{conclusion}).

\subsubsection{Topology model}
The simulated infrastructure represents a hierarchical smart city vehicular surveillance deployment organized into three tiers. At the top, the cloud tier (Node 0) is modeled as a central server (50,000 MIPS, 128 GB RAM, and GPU support) with a fixed 100 ms propagation delay to the gateway, reflecting typical wide-area network round-trip times. The gateway tier (Node 1) serves as an aggregation point with moderate resources (1,000 MIPS, 16 GB RAM) that bridges the cloud and the distributed edge. At the bottom, the edge tier (Nodes 2 through $N$) comprises heterogeneous nodes representing patrol vehicles randomly placed across a 1 km $\times$ 1 km urban area under a 5G geometric proximity model with a 300 m coverage radius.

To reflect real-world diversity, edge nodes are distributed as follows: 25\% NVIDIA Jetson-class (10,000 MIPS, GPU), 25\% Hailo-accelerator-equipped (10,000 MIPS, GPU), and 50\% Raspberry Pi-class (1,000 MIPS, CPU-only). Network links follow a 5G model with a 300 m coverage radius. As a result, approximately half of the edge nodes are GPU-capable, creating the kind of heterogeneity that makes hardware-aware allocation essential. Network links between edge nodes and the gateway carry propagation delays computed as the base 5G latency plus 2 ms km$^{-1}$, proportional to physical distance. Peer-to-peer links between geographically proximate edge nodes (within 300 m) carry half the base latency plus the distance-dependent term, modeling direct vehicular 5G communication.

\subsubsection{Application model}
The LPR pipeline serves as a stress-inducing workload for heterogeneous orchestration, transitioning from GPU-intensive inference (Vehicle and Plate Detection) to lightweight CPU-bound tasks (OCR and Notification). This worst-case scenario for hardware-aware allocation forces the orchestrator to satisfy conflicting resource affinities within a single execution chain. A policy that routes early stages correctly to GPU-capable nodes must simultaneously redirect later stages to CPU-sufficient nodes, forcing genuinely heterogeneous decisions at every pipeline invocation. This architectural characteristic makes LPR an ideal stress test for the \textit{Allocator}'s multi-criteria logic and justifies its use as the sole evaluative workload in this study.

The pipeline is modeled as a directed acyclic graph comprising four sequential stages, each calibrated to published performance profiles for the target hardware classes. In the first stage, Vehicle Detection receives 25 KB H.264-encoded key-frames, corresponding to a 6 Mbps stream subsampled to 1 FPS for LPR analysis~\cite{ammar2023multi}, and isolates vehicle regions using a YOLO-based detection model parameterized with $150 \times 10^6$ instructions, equivalent to approximately 15 ms of inference on a NVIDIA Jetson Xavier NX (21 TOPS, TensorRT)~\cite{kang2022evaluation}. 

The second stage, Plate Detection, processes 9 KB vehicle region-of-interest crops to localize license plate bounding boxes, also requiring $150 \times 10^6$ instructions. The third stage, OCR, applies character segmentation and classification to 5 KB cropped plate images using a lightweight convolutional recurrent neural network (CRNN) model parameterized with $10 \times 10^6$ instructions, corresponding to approximately 1 ms on GPU-enabled nodes. 

The final Notification stage transmits 100 B OCR results to the central cloud for logging and alert generation. A selectivity threshold of 1.0 is applied uniformly across all service modules, ensuring that every input frame propagates through all subsequent stages and thereby establishing a worst-case load scenario for the allocation strategies.

\subsubsection{Workload and network models}

Camera streams generate workload according to a Poisson process with arrival rate $\lambda = 1.0$ events/s$^{-1}$ per camera, corresponding to a 1 FPS key-frame submission rate for LPR analysis. This rate reflects the common operational practice in vehicular surveillance systems, where cameras record locally at full frame rate (e.g., 30 FPS) but forward only selected key-frames for inference to limit network and computational load~\cite{ammar2023multi}. The number of cameras per edge node is treated as an experimental variable with three levels (1, 2, or 4), enabling evaluation under light, moderate, and high workload conditions, respectively. At the highest intensity (four cameras across 100 edge nodes), the system must process up to 400 FPS, placing substantial stress on GPU-capable nodes and demanding accurate allocation decisions to prevent saturation of NVIDIA Jetson and Hailo devices while Raspberry Pi nodes (CPU-only) remain ineligible for GPU-accelerated stages. Notification sink modules are deployed on all edge nodes, allowing results to be consumed locally without requiring a round-trip to the cloud.

Three 5G network profiles, derived from 3GPP Release 16 specifications,\footnote{\url{https://www.3gpp.org/specifications-technologies/releases/release-16}} are evaluated to capture a range of operational conditions (see Table~\ref{tab:network_conditions}). We considered an ideal scenario (1 ms latency, 1,000 Mbps bandwidth) representing optimal mmWave 5G conditions, a typical urban scenario (5 ms, 500 Mbps) reflecting mid-band deployment, and a congested scenario (10 ms, 100 Mbps) modeling peak-hour network degradation.

\begin{table}[ht]
\centering
\small
\caption{5G network parameters}
\label{tab:network_conditions}
\begin{tabular}{lccc}
\hline
Scenario & Bandwidth (Mbps) & Latency (ms) & Jitter (ms) \\
\hline
Ideal (low latency) & 1000 & 1 & 0.5 \\
Typical (urban) & 500 & 5 & 1.0 \\
Congested & 100 & 10 & 2.0 \\
\hline
\end{tabular}
\end{table}

\subsubsection{Allocation Strategies}
We compared five allocation strategies, spanning two MCDM methods and three conventional heuristics. VIKOR, the default strategy adopted in CRAWO, computes a compromise ranking that jointly minimizes the global weighted utility measure $S$ and the maximum individual regret $R$ under three decision criteria: inference throughput in MIPS as a benefit criterion (40\% weight), network latency as a cost criterion (35\%), and current node queue length as a cost criterion (25\%), with the balance parameter set to $\nu = 0.5$. TOPSIS is included as an MCDM baseline that applies the same criteria and weights but selects the alternative closest to the ideal solution and farthest from the anti-ideal in the normalized criterion space.

Three heuristic baselines provide reference points of increasing sophistication. Random selection assigns each task uniformly at random among feasible nodes without inspecting the runtime state, serving as the simplest possible baseline. Round-Robin distributes tasks cyclically across all feasible nodes, implicitly assuming uniform processing capabilities. Least-Loaded selects the node with the smallest current queue length at dispatch time, thereby optimizing for immediate resource availability but without accounting for hardware heterogeneity or network distance.

\subsection{Experimental design}
\label{sec:exp_design}
The evaluation follows a full factorial design with two related test campaigns, summarized in Table~\ref{tab:exp_parameters}.

\begin{table}[ht]
\centering
\small
\caption{Experimental parameters and scenario configurations.}
\label{tab:exp_parameters}
\renewcommand{\arraystretch}{1.15}
\begin{tabular}{>{\raggedright}p{2cm}>{\raggedright}p{8cm}c}
\multicolumn{3}{l}{\textbf{Scenario 1: Strategy Comparison (RQ1--RQ2)}} \\ 
\hline
Parameter & Values & Combinations \\
\hline
Edge nodes         & 10, 50, 100 & 3 \\
Cameras per node   & 1, 2, 4 & 3 \\
Strategies         & TOPSIS, VIKOR, Random, Round-Robin, Least-Loaded & 5 \\
Network conditions & 1 ms/1 Gbps, 5 ms/500 Mbps, 10 ms/100 Mbps & 3 \\
Seeds              & 1--10 & 10 \\
\hline
\multicolumn{3}{r}{Total runs = 1,350} \\
\multicolumn{3}{l}{\textbf{Scenario 2: Deployment Architecture (RQ3)}} \\ 
\hline
Parameter & Values & Combinations \\
\hline
Edge nodes         & 100 (fixed) & 1 \\
Cameras per node   & 2 (fixed) & 1 \\
Deployment         & distributed/VIKOR, cloud\_only/Random & 2 \\
Network conditions & 5ms/500Mbps (fixed) & 1 \\
Seeds              & 1--10 & 10 \\
\hline
\multicolumn{3}{r}{Total runs = 20} \\
\end{tabular}
\end{table}

\textit{Scenario 1 -- Strategy Comparison} crosses all five allocation strategies against three deployment scales (10, 50, and 100 edge nodes), three workload intensities (1, 2, and 4 cameras per node), and three 5G network profiles. With ten independent random seeds, this yields a total of 1,350 simulation runs. At one camera per node, GPU utilization remains near 12\%, making most strategies perform similarly. However, with 4 cameras per node, NVIDIA Jetson and Hailo devices approach saturation, while Raspberry Pi nodes remain ineligible for GPU stages, forcing the allocator to actively differentiate between heterogeneous hardware capabilities.

\textit{Scenario 2 -- Deployment Architecture} isolates the impact of processing placement. It compares edge-distributed execution (under the VIKOR strategy) against a cloud-only baseline at a fixed scale of 100 nodes and two cameras per node under typical 5G conditions (5 ms, 500 Mbps). Replicated across ten seeds, this adds 20 simulation runs. To ensure consistent data collection, each simulation runs for 600 s, with the first 120 s as a warm-up period to bring the system to steady state before performance measurements are recorded.

\subsection{Simulation implementation}
\label{sec:sim_impl}
We implemented the simulation by extending the five core abstractions provided by the YAFS framework~\cite{lera2019yafs}. The \textit{Topology Abstraction} models the physical infrastructure as a weighted graph, where nodes carry computational attributes (processing capacity, memory, GPU availability) and edges carry network parameters (bandwidth and propagation delay). The \textit{Application Abstraction} defines the LPR pipeline as a directed acyclic graph of service modules connected by typed \textit{Message} objects, each parameterized with an instruction count and a payload size in bytes.

Two placement policies govern the initial deployment of service modules onto topology nodes. A \textit{Distributed Policy} that registers all pipeline stages on every node, delegating all runtime routing decisions to the selection strategy, and a \textit{Cloud-Only Policy} restricts all modules exclusively to the cloud node, thereby establishing the centralized baseline. A custom \textit{Population Policy} generates workload by attaching a configurable number of camera sources per edge node (1, 2, or 4), each emitting key frames according to an exponential inter-arrival distribution with rate $\lambda = 1.0$~events~s$^{-1}$, and by placing notification sinks at all edge nodes to enable local result collection. Finally, the \textit{Selection Policy} is invoked at runtime for every in-flight message to determine the target service instance and the associated routing path. All five allocation strategies are implemented as \textit{Selection} subclasses, each overriding the \texttt{get\_path} method.

\subsection{Results}
\label{sec:results}
This section presents the experimental results derived from the two test campaigns, directly addressing the three research questions. Section~\ref{sec:results_test1} details the results for \textit{Scenario 1 -- Strategy Comparison}, which evaluate the efficacy and decision quality (RQ1) as well as the scalability (RQ2) of the allocation strategies across varying scales, workloads, and 5G network profiles. Section~\ref{sec:results_test2} presents the results for \textit{Scenario 2 -- Deployment Architecture}, addressing the architectural trade-offs (RQ3) by comparing edge-distributed execution with a cloud-only baseline.

\subsubsection{Scenario 1 results}
\label{sec:results_test1}
Table~\ref{tab:latency_results} reports the mean end-to-end pipeline latency (ms) by strategy, deployment scale, and workload intensity across three 5G network profiles, averaged over ten seeds. In turn, Table~\ref{tab:utilization_results} presents the corresponding load-balancing coefficient of variation (CV\,$=\sigma/\mu$) and pipeline failure rate for each strategy, deployment scale, and workload intensity across three 5G network profiles, averaged over ten seeds.

\begin{landscape}
\begin{table}[ht]
\centering
\caption{Mean end-to-end latency across all experimental conditions}
\label{tab:latency_results}
\small
\setlength{\tabcolsep}{6pt}
\begin{tabular}{l rrr rrr rrr}
\hline
\multirow{2}{*}{Strategy} & \multicolumn{3}{c}{10 nodes} & \multicolumn{3}{c}{50 nodes} & \multicolumn{3}{c}{100 nodes} \\
\cline{2-10}
& 1 cam & 2 cam & 4 cam & 1 cam & 2 cam & 4 cam & 1 cam & 2 cam & 4 cam \\
\hline
\multicolumn{10}{l}{\textit{Ideal --- 1 ms / 1,000 Mbps}}\\[2pt]
TOPSIS       &  31.3 &  31.6 &   32.4 &  31.3 &  31.7 &    32.5 &  31.4 &  31.7 &    32.6 \\
VIKOR        &  31.3 &  31.6 &   32.4 &  31.3 &  31.7 &    32.5 &  31.4 &  31.7 &    32.6 \\
Random       & 202.8 & 219.7 &  378.2 & 195.3 & 214.3 & 1,788.7 & 196.4 & 215.9 & 1{,}166.4 \\
Round-Robin  & 172.0 & 183.2 &  229.6 & 181.3 & 181.8 & 1,665.6 & 182.9 & 183.2 & 1,046.0 \\
Least-Loaded & 200.5 & 210.4 & 2,966.7 & 191.5 & 198.8 & 2,957.2 & 192.8 & 199.0 & 1,752.6 \\
\hline
\multicolumn{10}{l}{\textit{Typical --- 5 ms / 500 Mbps}}\\[2pt]
TOPSIS       &  38.6 &  38.7 &   39.5 &  31.4 &  31.9 &    32.9 &  31.4 &  31.9 &    33.0 \\
VIKOR        &  31.4 &  31.9 &   32.9 &  31.4 &  31.9 &    32.9 &  31.4 &  31.9 &    33.0 \\
Random       & 219.6 & 236.5 &  388.3 & 212.4 & 231.7 & 1,731.7 & 213.7 & 233.1 & 1,188.8 \\
Round-Robin  & 184.3 & 197.9 &  247.2 & 198.5 & 199.0 & 1,682.4 & 199.8 & 200.2 & 1,061.1 \\
Least-Loaded & 217.8 & 226.2 & 3,112.8 & 209.4 & 216.1 & 3,041.2 & 209.4 & 216.7 & 1,756.9 \\
\hline
\multicolumn{10}{l}{\textit{Congested --- 10 ms / 100 Mbps}} \\[2pt]
TOPSIS       & 110.1 & 1,563.1$^\dagger$ & 2,370.2$^\dagger$ & 7,912.4$^\dagger$ & 7,417.1$^\dagger$ & 5,660.8$^\dagger$ &  31.6 &  32.1 &  33.6 \\
VIKOR        &  42.6 &    42.8 &    42.9 &    31.5 &    32.1 &    33.4 &  31.6 &  32.1 &  33.6 \\
Random       & 239.9 & 256.5 &  424.7 & 234.4 & 254.0 & 1,815.3 & 235.3 & 255.0 & 1,213.6 \\
Round-Robin  & 200.0 & 216.6 &  265.9 & 220.1 & 220.9 & 1,703.6 & 221.3 & 221.7 & 1,088.9 \\
Least-Loaded & 238.1 & 248.6 & 2,748.3 & 230.9 & 239.1 & 2,992.4 & 231.1 & 238.5 & 1,756.4 \\
\hline
\multicolumn{10}{l}{\scriptsize$^\dagger$ denotes TOPSIS routing instability under congestion}
\end{tabular}
\end{table}
\end{landscape}

\begingroup
\small
\setlength{\tabcolsep}{3pt}
\begin{longtable}{l rrr rrr rrr}
\caption{Load coefficient of variation (CV) and pipeline failure rates}
\label{tab:utilization_results} \\
\hline
\multirow{2}{*}{Strategy} & \multicolumn{3}{c}{10 nodes} & \multicolumn{3}{c}{50 nodes} & \multicolumn{3}{c}{100 nodes} \\
\cline{2-10}
& 1 cam & 2 cam & 4 cam & 1 cam & 2 cam & 4 cam & 1 cam & 2 cam & 4 cam \\
\hline
\endfirsthead

\multicolumn{10}{c}
{{\tablename\ \thetable{} (continued): Load coefficient of variation (CV) and pipeline failure rates}} \\
\hline
\multirow{2}{*}{Strategy} & \multicolumn{3}{c}{10 nodes} & \multicolumn{3}{c}{50 nodes} & \multicolumn{3}{c}{100 nodes} \\
\cline{2-10}
& 1 cam & 2 cam & 4 cam & 1 cam & 2 cam & 4 cam & 1 cam & 2 cam & 4 cam\\
\hline
\endhead

\hline
\multicolumn{10}{r}{{\scriptsize Continued on next page}} \\
\endfoot

\hline
\endlastfoot

\multicolumn{10}{l}{\textit{CV --- Ideal (1 ms / 1,000 Mbps)}} \\[2pt]
TOPSIS        & 0.541 & 0.527 & 0.481 & 0.517 & 0.491 & 0.448 & 0.540 & 0.510 & 0.454 \\
VIKOR         & 0.541 & 0.527 & 0.481 & 0.517 & 0.491 & 0.448 & 0.540 & 0.510 & 0.454 \\
Random        & 0.031 & 0.020 & 0.015 & 0.026 & 0.019 & 0.021 & 0.027 & 0.019 & 0.020 \\
Round-Robin   & 0.001 & 0.000 & 0.000 & 0.001 & 0.000 & 0.017 & 0.001 & 0.000 & 0.014 \\
Least-Loaded  & 0.031 & 0.042 & 0.121 & 0.028 & 0.044 & 0.145 & 0.028 & 0.044 & 0.151 \\
\hline
\multicolumn{10}{l}{\textit{CV --- Typical (5 ms / 500 Mbps)}} \\[2pt]
TOPSIS        & 0.437 & 0.433 & 0.400 & 0.517 & 0.491 & 0.448 & 0.540 & 0.510 & 0.454 \\
VIKOR         & 0.541 & 0.527 & 0.481 & 0.517 & 0.491 & 0.448 & 0.540 & 0.510 & 0.454 \\
Random        & 0.026 & 0.020 & 0.015 & 0.028 & 0.018 & 0.022 & 0.026 & 0.019 & 0.020 \\
Round-Robin   & 0.001 & 0.000 & 0.000 & 0.001 & 0.000 & 0.017 & 0.001 & 0.000 & 0.014 \\
Least-Loaded  & 0.025 & 0.044 & 0.119 & 0.027 & 0.044 & 0.144 & 0.029 & 0.044 & 0.150 \\
\hline
\multicolumn{10}{l}{\textit{CV --- Congested (10 ms / 100 Mbps)}} \\[2pt]
TOPSIS        & 0.546 & 0.469 & 0.408 & 1.847$^\dagger$ & 1.578$^\dagger$ & 1.406$^\dagger$ & 0.540 & 0.510 & 0.455 \\
VIKOR         & 0.448 & 0.444 & 0.381 & 0.517 & 0.491 & 0.449 & 0.540 & 0.510 & 0.455 \\
Random        & 0.030 & 0.019 & 0.014 & 0.025 & 0.019 & 0.021 & 0.025 & 0.019 & 0.020 \\
Round-Robin   & 0.001 & 0.000 & 0.000 & 0.001 & 0.000 & 0.017 & 0.001 & 0.000 & 0.014 \\
Least-Loaded  & 0.030 & 0.041 & 0.120 & 0.030 & 0.044 & 0.144 & 0.029 & 0.045 & 0.150 \\
\hline
\multicolumn{10}{l}{\textit{Failure rate (\%) --- Ideal (1 ms / 1,000 Mbps)}} \\[2pt]
TOPSIS        & 0.00 & 0.00 & 0.00 & 0.01 & 0.00 & 0.00 & 0.00 & 0.00 & 0.00 \\
VIKOR         & 0.00 & 0.00 & 0.00 & 0.01 & 0.00 & 0.00 & 0.00 & 0.00 & 0.00 \\
Random        & 0.00 & 0.01 & 0.01 & 0.02 & 0.01 & 0.45 & 0.01 & 0.01 & 0.29 \\
Round-Robin   & 0.00 & 0.00 & 0.00 & 0.01 & 0.01 & 0.47 & 0.01 & 0.01 & 0.30 \\
Least-Loaded  & 0.00 & 0.00 & 0.73 & 0.02 & 0.01 & 1.01 & 0.02 & 0.01 & 0.59 \\
\hline
\multicolumn{10}{l}{\textit{Failure rate (\%) --- Typical (5 ms / 500 Mbps)}} \\[2pt]
TOPSIS        & 0.00 & 0.00 & 0.00 & 0.01 & 0.00 & 0.00 & 0.00 & 0.00 & 0.00 \\
VIKOR         & 0.00 & 0.00 & 0.00 & 0.01 & 0.00 & 0.00 & 0.00 & 0.00 & 0.00 \\
Random        & 0.00 & 0.01 & 0.01 & 0.01 & 0.02 & 0.43 & 0.01 & 0.01 & 0.29 \\
Round-Robin   & 0.00 & 0.00 & 0.00 & 0.01 & 0.01 & 0.46 & 0.01 & 0.01 & 0.30 \\
Least-Loaded  & 0.00 & 0.01 & 0.71 & 0.01 & 0.01 & 1.01 & 0.01 & 0.01 & 0.58 \\
\hline
\multicolumn{10}{l}{\textit{Failure rate (\%) --- Congested (10 ms / 100 Mbps)}} \\[2pt]
TOPSIS        & 0.00 & 0.64$^\dagger$ & 1.81$^\dagger$ & 8.42$^\dagger$ & 13.70$^\dagger$ & 20.28$^\dagger$ & 0.00 & 0.00 & 0.00 \\
VIKOR         & 0.00 & 0.00 & 0.00 & 0.00 & 0.00 & 0.00 & 0.00 & 0.00 & 0.00 \\
Random        & 0.00 & 0.00 & 0.03 & 0.01 & 0.02 & 0.44 & 0.02 & 0.00 & 0.28 \\
Round-Robin   & 0.00 & 0.00 & 0.00 & 0.01 & 0.02 & 0.46 & 0.02 & 0.00 & 0.30 \\
Least-Loaded  & 0.00 & 0.00 & 0.55 & 0.01 & 0.02 & 0.97 & 0.02 & 0.01 & 0.58 \\
\hline
\multicolumn{10}{l}{\scriptsize$^\dagger$ denotes TOPSIS routing instability under congestion}
\end{longtable}
\endgroup

Three principal observations emerge from the analysis of these two tables. First, both MCDM methods (VIKOR and TOPSIS) consistently achieve substantially lower latencies than the three heuristic baselines across all tested configurations. Under typical 5G conditions at 100 nodes and four cameras per node, VIKOR and TOPSIS deliver mean latencies of 33 ms, compared to 1,061 ms for Round-Robin (a 96.9\% reduction), 1,189 ms for Random (97.2\% reduction), and 1,757 ms for Least-Loaded (98.1\% reduction).

Second, VIKOR and TOPSIS perform identically at 50 and 100 nodes under ideal and typical 5G profiles. Under these favorable conditions, both MCDM strategies exhibit CV values of approximately 0.44--0.54 (see Table~\ref{tab:utilization_results}), which reflect intentional GPU-preferential routing rather than load imbalance. However, under the congested profile (10 ms, 100 Mbps), TOPSIS exhibits routing instability at 10 and 50 nodes, with latencies escalating to 7,912 ms (50 nodes, one camera), CV exceeding 1.0, and failure rates reaching 20.28\% (50 nodes, four cameras). In contrast, VIKOR remains stable across all conditions, maintaining latencies below 43 ms and failure rates lower than 0.005\%. This confirms VIKOR's suitability for dynamic edge environments where criteria are inherently conflicting.

\begin{figure}[ht]
    \centering
    \includegraphics[width=\textwidth]{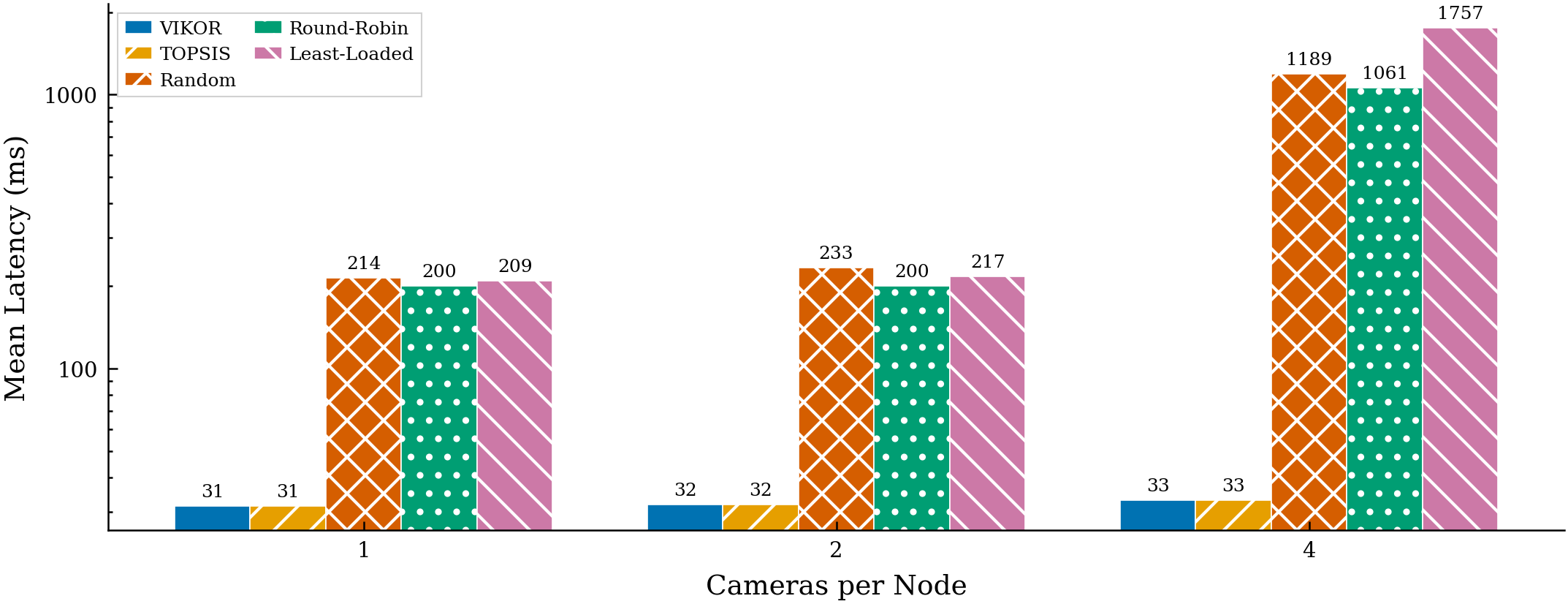}
    \caption{Mean end-to-end latency (ms, log scale) as a function of workload intensity (cameras per node) for all five allocation strategies at 100 edge nodes under typical 5G conditions (5 ms, 500 Mbps).}
    \label{fig:workload_intensity}
\end{figure}

Third, the performance gap between MCDM and heuristics widens as workload intensity increases (see \figurename~\ref{fig:workload_intensity}). This reflects the critical role of hardware-aware allocation as GPU-capable nodes approach saturation. Heuristic baselines, lacking hardware awareness, increasingly misroute tasks to CPU-only Raspberry Pi nodes, causing cascading queue buildup and system-wide latency spikes.

\subsubsection{Scenario 2 results}
\label{sec:results_test2}
\begin{figure}[ht]
    \centering
    \includegraphics[width=\textwidth]{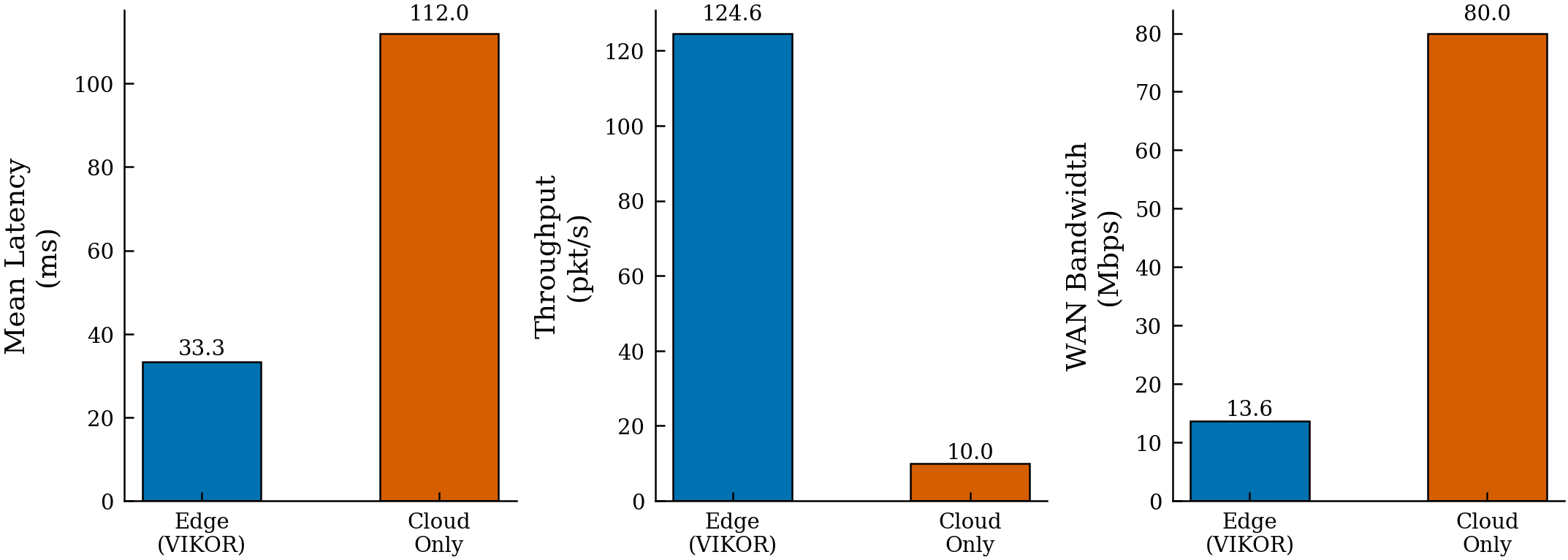}
    \caption{Performance comparison of edge-distributed (VIKOR) versus cloud-only deployment across three metrics at 100 edge nodes and two cameras per node under typical 5G conditions (5 ms / 500 Mbps): mean end-to-end latency, pipeline throughput, and network bandwidth consumption.}
    \label{fig:edge_vs_cloud}
\end{figure}

Table~\ref{tab:deployment_comparison} and \figurename~\ref{fig:edge_vs_cloud} compare edge-distributed execution under VIKOR against the cloud-only baseline at 100 edge nodes with two cameras per node under typical 5G conditions. All throughput and load-served values are measured over the steady-state window (120--600 s) against an offered load of 200 FPS (100 nodes $\times$ 2 cameras $\times$ 1 FPS). The edge-distributed deployment achieves a mean end-to-end latency of 31.9 ms with a P95 of 37.8 ms, compared to 112.0 ms and 112.4 ms for the cloud-only configuration.

\begin{table}[ht]
\centering
\small
\caption{Edge-distributed (VIKOR) vs. cloud-only deployment at 100 edge nodes, two cameras per node, and typical 5G (5 ms / 500 Mbps), averaged over 10 seeds.}
\label{tab:deployment_comparison}
\begin{tabular}{lcc}
\hline
Metric & Edge-distributed (VIKOR) & Cloud-only \\
\hline
Mean E2E latency (ms)   &  31.9  & 112.0 \\
P95 latency (ms)        &  37.8  & 112.4 \\
Throughput (pipeline/s) & 199.9  &  10.0 \\
Load served (\%)        &  99.9  &   5.0 \\
Bandwidth usage (Mbps)  &  21.2  &  80.0 \\
\hline
Latency reduction (\%)  & \multicolumn{2}{l}{71.5\% (edge vs.\ cloud)} \\
Throughput gain         & \multicolumn{2}{l}{$\approx$20$\times$ (edge vs.\ cloud)} \\
\hline
\end{tabular}
\end{table}

More critically, the edge deployment sustains a throughput of 199.9 pipelines/s, serving 99.9\% of the offered load, whereas the cloud-only server (despite its 50,000 MIPS capacity) reaches compute saturation at approximately 161 FPS and processes only 10.0 pipelines/s (5.0\% of the load), leaving 95\% of submitted frames unprocessed in the queue. Network bandwidth consumption drops from 80.0 Mbps under the cloud-only configuration to 21.2 Mbps under edge-distributed execution, a 74\% reduction.

\subsection{Discussion}
\label{sec:discussion}

This section discusses the CRAWO's experimental results, focusing on the efficacy of the VIKOR allocation strategy (RQ1), its scalability across varying node counts and 5G conditions (RQ2), and the architectural advantages of edge execution over the cloud in terms of latency and network bandwidth (RQ3).

\subsubsection{Strategy efficacy and decision quality (RQ1)}
Within the evaluated LPR pipeline and 5G topology, the results indicate that the MCDM-based allocation significantly outperforms conventional heuristics. Under typical 5G conditions (5 ms, 500 Mbps) at 100 edge nodes, VIKOR reduces mean end-to-end latency by 85\% compared to Random selection (31.4 ms vs. 213.7 ms at one camera per node) and by 85--98\%compared to Least-Loaded across all workload intensities. The performance gap is most evident at four cameras per node, where VIKOR achieves 33.0 ms compared to 1,757 ms for Least-Loaded, representing a 53-fold improvement. This widening advantage reflects the critical role of hardware-aware allocation. As GPU-capable devices (NVIDIA Jetson and Hailo) approach saturation, any allocator that ignores hardware heterogeneity misroutes tasks to CPU-only Raspberry Pi nodes. This triggers a cascading buildup of queues and order-of-magnitude latency degradation.

The comparison between the two MCDM algorithms also reveals an important distinction in robustness:

\begin{itemize}[noitemsep]
  \item \textit{Convergence in ideal conditions:} TOPSIS and VIKOR deliver nearly identical latencies at 50 and 100 nodes under favorable network conditions, confirming both methods converge to near-optimal GPU-preferential routing.
  \item \textit{Sensitivity to topology scale:} VIKOR achieves sub-32 ms latency even at ten nodes, whereas TOPSIS lags (38.6 ms). This suggests VIKOR's compromise-regret formulation is less sensitive to queue-length fluctuations in smaller topologies.
  \item \textit{Resilience to congestion:} Under the congested profile (10 ms, 100 Mbps), TOPSIS exhibits instability, with latencies escalating to 7,912 ms and failure rates reaching 20.28\%, as shown in Table~\ref{tab:utilization_results}. This occurs because elevated network delays inflate the queue-state data, causing TOPSIS to over-concentrate load on single nodes. VIKOR's regret-minimization component acts as a safeguard against such concentration, ensuring that no single criterion (such as a stale queue metric) dominates the decision.
\end{itemize}

\subsubsection{Scalability (RQ2)}
VIKOR exhibits good scale invariance, maintaining a mean latency below 33 ms across 10 to 100 nodes regardless of workload or network conditions. This consistency suggests that the compromise-regret formulation effectively selects GPU nodes even in constrained ten-node deployments with a low ratio of GPU-capable to CPU-only nodes. In contrast, TOPSIS requires a higher node density (50 or more) to achieve comparable performance under typical 5G conditions.

The heuristic baselines exhibit substantially higher sensitivity to both scale and workload intensity. Least-Loaded, for example, incurs latencies of 3,113 ms and 3,041 ms at 10 and 50 nodes, respectively, with four cameras per node before improving to 1,757 ms at 100 nodes, still two orders of magnitude above VIKOR (representing VIKOR latency reductions of 98.9\%, 98.9\%, and 98.1\% relative to Least-Loaded at 10, 50, and 100 nodes, respectively). This behavior highlights a limitation of queue-length-based heuristics in the heterogeneous environment tested: minimizing queue length without considering per-node processing capacity misroutes GPU-dependent tasks to CPU-only nodes. Across all 1,350 simulation runs in Scenario 1, VIKOR failure rates remain below 0.01\%, thereby indicating reliable execution within the tested scale range and network profiles.

The heuristic baselines are highly sensitive to both scale and workload. For instance, Least-Loaded incurs latencies exceeding 3,000 ms at smaller scales (10 and 50 nodes with four cameras per node) before improving slightly at 100 nodes, even though it remains two orders of magnitude slower than VIKOR. This behavior highlights the inherent limitation of queue-length heuristics in the tested heterogeneous environment: minimizing queue length without accounting for processing capacity misroutes GPU-dependent tasks to CPU-only nodes. Across all 1,350 simulation runs in Scenario 1, VIKOR failure rates remained below 0.01\%, proving its reliability across the tested scale range and network profiles.

\subsubsection{Deployment architecture (RQ3)}
Scenario 2 quantifies the benefits of edge-distributed processing. VIKOR-based edge execution reduces mean end-to-end latency by 71.5\% relative to cloud-only deployment (31.9 ms vs. 112.0 ms) and improves throughput by approximately 20 times (199.9 vs. 10.0 pipelines/s), serving 99.9\% of the offered 200 FPS load compared to only 5.0\% for the cloud-only configuration (see Table~\ref{tab:deployment_comparison}). The cloud server, despite its 50,000 MIPS capacity, reaches compute saturation at 161 FPS, leaving 95\% of frames unprocessed within the 480 s measurement window.

Beyond latency and throughput, CRAWO reduces network bandwidth consumption by 74\%. Because only 100-byte notification messages traverse the wide-area link (rather than 25 KB full frames), the distributed architecture significantly reduces infrastructure costs and preserves network capacity for concurrent smart city services.

\section{Conclusion}
\label{conclusion}
The expansion of smart city applications, particularly real-time AI video processing, exposes the critical limitations of cloud-only architectures regarding prohibitive latency and significant pressure on network bandwidth. While shifting computation to the edge mitigates these issues, the inherent hardware heterogeneity of these environments makes static deployments highly inefficient. Consequently, AI-based pipelines executed across diverse edge infrastructures require an intelligent, adaptive orchestration approach.

To address this challenge, this work introduced the CRAWO framework, designed to manage distributed AI pipelines across heterogeneous edge clusters. Operating alongside Kubernetes, CRAWO employs a control-loop model divided into three functional planes. The \textit{Control Plane} is responsible for intelligent decision-making, the \textit{Data Plane} handles domain modeling using Kubernetes CRDs, and the \textit{Execution Plane} ensures continuous state reconciliation. By decoupling orchestration logic from execution mechanisms, CRAWO treats processing pipelines as cohesive units and dynamically places stages based on real-time runtime conditions.

A fundamental contribution of the CRAWO architecture is its hardware-aware allocator, which favors an MCDM approach over rigid scheduling rules. Instead of relying on simplistic heuristics such as Random or Round-Robin, CRAWO utilizes the VIKOR method to calculate a compromise ranking for task placement. This strategy balances conflicting goals by maximizing inference throughput while minimizing network latency and node load, ensuring that intensive tasks are routed to appropriate GPU-enabled nodes.

We evaluated CRAWO through discrete-event simulation using the YAFS framework, modeling a multi-stage LPR pipeline deployed on mobile patrol vehicles under varying 5G network conditions. The results demonstrated the superiority of the VIKOR-based approach for:

\begin{itemize}[noitemsep]
  \item Performance: end-to-end latency was reduced by up to 98\% under high workload intensities compared to traditional heuristics.
  \item Robustness: unlike the TOPSIS baseline, which exhibited severe routing instability and 20.28\% failure rates in congested networks, VIKOR maintained stable routing with latencies consistently below 43 ms.
  \item Architectural efficiency: edge-distributed execution reduced mean latency by 71.5\%, boosted throughput by 20 times, and lowered network bandwidth consumption by 74\% compared to a cloud-only model.
\end{itemize}

Despite these results, this work has limitations that should be acknowledged. The evaluation relied on discrete-event simulation, while real-world deployments may introduce additional variables not captured by the simulation model, such as hardware failures, operating system scheduling interference, and dynamic 5G handoffs. Additionally, the study focused on a single LPR use case in a vehicular surveillance context, and we fixed the VIKOR method's allocation criterion weights. Generalization to other smart city pipelines or edge topologies requires further investigation.

Future work will focus on four directions. First, experimental validation on physical heterogeneous hardware clusters, including NVIDIA Jetson and Hailo-equipped nodes, will complement simulation results presented in this work. Next, we will extend VIKOR weight configuration to support adaptive or learned parameterization based on observed runtime performance, reducing the need for manual tuning. Subsequently, we will evaluate CRAWO on additional AI pipeline use cases, such as anomaly detection and environmental monitoring, to assess its generalizability across diverse smart city applications. Finally, we plan to replace the current relational database used for allocation state persistence with the \textit{etcd} key-value store native to Kubernetes. This migration will expose CRAWO’s state directly through the Kubernetes API, enabling atomic watch-based notifications and seamless integration with the CRD reconciliation loop, thus ultimately reducing operational complexity and aligning the framework with modern cloud-native models.

\section*{CRediT authorship contribution statement}
\textbf{Eugênio Santos:} Investigation, Software, Formal analysis, Visualization, Writing – original draft. 
\textbf{Daniel Maia:} Investigation, Software, Formal analysis, Visualization, Writing – original draft. 
\textbf{Stefano Loss:} Investigation, Software, Supervision, Writing – original draft, Writing – review \& editing. 
\textbf{José Manoel Silva:} Investigation, Software. 
\textbf{Aluizio Rocha Neto:} Methodology, Supervision, Validation, Writing – review \& editing.
\textbf{Thais Batista:} Funding acquisition, Project administration, Validation, Writing – review \& editing. 
\textbf{Everton Cavalcante:} Validation, Writing – review \& editing.
\textbf{Nélio Cacho:} Validation, Writing – review \& editing.
\textbf{Eduardo Nogueira:} Validation, Writing – review \& editing. 
\textbf{Daniel Araújo:} Methodology, Supervision, Validation, Writing – review \& editing.
\textbf{Frederico Lopes:} Funding acquisition, Project administration, Validation, Writing – review \& editing.

\section*{Funding sources}
This work is supported by the SPICI project, funded by FINEP (grant 2827/22), and the INCT  Intelligent Communications Networks and the Internet of Things - ICoNIoT, funded by CNPq (grant 405940/2022-0) and CAPES (grant 88887.954253/2024-00).

\section*{Declaration of competing interest}
The authors declare that they have no known competing financial interests or personal relationships that could have appeared to influence the work reported in this paper.

\bibliographystyle{elsarticle/elsarticle-num}
\bibliography{references}

@Article{cheng2017fogflow,
  author  = {Bin Cheng and G{\"u}rkan Solmaz and Flavio Cirillo and Ern{\"o} Kovacs and Kazuyuki Terasawa and Atsushi Kitazawa},
  journal = {IEEE Internet of Things Journal},
  title   = {{FogFlow}: Easy programming of {IoT} services over cloud and edges for smart cities},
  year    = {2017},
  month   = apr,
  number  = {2},
  pages   = {696--707},
  volume  = {5},
  doi     = {10.1109/JIOT.2017.2747214},
}

@Article{bohm2022cloud,
  author  = {Sebastian B{\"o}hm and Guido Wirtz},
  journal = {EAI Endorsed Transactions on Smart Cities},
  title   = {Cloud-edge orchestration for smart cities: A review of {Kubernetes}-based orchestration architectures},
  year    = {2022},
  month   = may,
  number  = {18},
  volume  = {6},
  doi     = {10.4108/eetsc.v6i18.1197},
}

@Book{papathanasiou2018topsis,
  author    = {Jason Papathanasiou and Nikolaos Ploskas},
  publisher = {Springer},
  title     = {Multiple Criteria Decision Aid: Methods, Examples and Python Implementations},
  year      = {2018},
  address   = {Cham, Switzerland},
  doi       = {10.1007/978-3-319-91648-4},
}

@Article{sonnara2025efficient,
  author  = {Fedi Sonnara and Hamadi Chihaoui and Fethi Filali},
  journal = {Journal of Real-Time Image Processing},
  title   = {Efficient real-time license plate recognition using deep learning on edge devices},
  year    = {2025},
  number  = {4},
  pages   = {159},
  volume  = {22},
  doi     = {10.1007/s11554-025-01738-3},
}

@Article{ruiu2025continuum,
  author  = {Pietro Ruiu and Andrea Lagorio and Claudio Rubattu and Matteo Anedda and Michele Sanna and Mauro Fadda},
  journal = {Future Internet},
  title   = {Edge-to-cloud continuum orchestrator based on heterogeneous nodes for urban traffic monitoring},
  year    = {2025},
  month   = dec,
  number  = {12},
  volume  = {17},
  doi     = {10.3390/fi17120574},
}

@Article{silva2022fogsurvey,
  author    = {Thiago Pereira da Silva and Thais Batista and Frederico Lopes and Aluizio Rocha Neto and Fl{\'a}via C. Delicato and Paulo F. Pires and Atslands R. da Rocha},
  journal   = {ACM Transactions on Internet Technology},
  title     = {Fog computing platforms for smart city applications: A survey},
  year      = {2022},
  month     = nov,
  number    = {4},
  volume    = {22},
  articleno = {96},
  doi       = {10.1145/3488585},
}

@Article{singh2023edgeai,
  author  = {Raghubir Singh and Sukhpal Singh Gill},
  journal = {Internet of Things and Cyber-Physical Systems},
  title   = {Edge {AI}: A survey},
  year    = {2023},
  pages   = {71--92},
  volume  = {3},
  doi     = {10.1016/j.iotcps.2023.02.004},
}

@PhdThesis{rochaneto2021thesis,
  author  = {Aluizio Ferreira {Rocha Neto}},
  school  = {Federal University of Rio Grande do Norte},
  title   = {Edge-distributed stream processing for video analytics in smart city applications},
  year    = {2021},
  address = {Natal, Brazil},
  url     = {https://repositorio.ufrn.br/handle/123456789/32743},
}

@Article{zeydan2022mcdm,
  author  = {Engin Zeydan and Josep Mangues-Bafalluy and Jorge Baranda and Ricardo Mart{\'i}nez and Luca Vettori},
  journal = {Journal of Network and Systems Management},
  title   = {A multi-criteria decision making approach for scaling and placement of virtual network functions},
  year    = {2022},
  number  = {2},
  pages   = {32},
  volume  = {30},
  doi     = {10.1007/s10922-022-09645-9},
}

@Article{lera2019yafs,
  author  = {Isaac Lera and Carlos Guerrero and Carlos Juiz},
  journal = {IEEE Access},
  title   = {{YAFS}: A Simulator for {IoT} Scenarios in fog computing},
  year    = {2019},
  pages   = {91745--91758},
  volume  = {7},
  doi     = {10.1109/ACCESS.2019.2927895},
}

@Article{ammar2023multi,
  author  = {Adel Ammar and Anis Koubaa and Wadii Boulila and Bilel Benjdira and Yasser Alhabashi},
  journal = {Sensors},
  title   = {A multi-stage deep-learning-based vehicle and license plate recognition system with real-time edge inference},
  year    = {2023},
  number  = {4},
  volume  = {23},
  doi     = {10.3390/s23042120},
}

@Article{kang2022evaluation,
  author  = {Pilsung Kang and Athip Somtham},
  journal = {Mathematics},
  title   = {An evaluation of modern accelerator-based edge devices for object detection applications},
  year    = {2022},
  number  = {22},
  volume  = {10},
  doi     = {10.3390/math10224299},
}

@Article{opricovic2004vikor,
  author  = {Serafim Opricovic and Gwo-Hshiung Tzeng},
  journal = {European Journal of Operational Research},
  title   = {Compromise Solution by MCDM Methods: A Comparative Analysis of VIKOR and TOPSIS},
  year    = {2004},
  number  = {2},
  pages   = {445--455},
  volume  = {156},
  doi     = {10.1016/S0377-2217(03)00020-1},
}

@InProceedings{loss2025framework,
  author    = {Loss, Stefano and Costa, Karine and Gurgel, Leonandro and Sales, Ravelly and Brito, Vicente and Batista, Thais and Cavalcante, Everton and Lopes, Frederico and Neto, Aluizio Rocha and Cacho, Nélio},
  booktitle = {2025 IEEE Latin Conference on IoT (LCIoT)},
  title     = {A framework for live situational awareness in stream-based {5G} applications},
  year      = {2025},
  pages     = {202--205},
  publisher = {IEEE},
  doi       = {10.1109/lciot64881.2025.11118567},
}

@InCollection{lira2025enhancing,
  author    = {Lira, Pedro and Costa, Karine and Loss, Stefano and Lima, Leonardo and Araújo, Daniel and Nogueira, Eduardo and Neto, Aluizio Rocha and Batista, Thais and Cacho, Nelio and Cavalcante, Everton and Lopes, Frederico},
  booktitle = {Intelligent Systems},
  publisher = {Springer, Cham},
  title     = {Enhancing situational awareness in public safety with frame-accumulated face recognition and distance-based evaluation},
  year      = {2026},
  address   = {Switzerland},
  editor    = {de Freitas, Rosiane and Furtado, Diego},
  pages     = {245--259},
  series    = {Lecture Notes in Computer Science},
  volume    = {16182},
  doi       = {10.1007/978-3-032-15993-9\_17},
}

@Article{rosmaninho2024edgecloud,
  author  = {Rosmaninho, Rodrigo and Raposo, Duarte and Rito, Pedro and Sargento, Susana},
  journal = {IEEE Transactions on Services Computing},
  title   = {Edge-cloud continuum orchestration of critical services: A smart-city approach},
  year    = {2025},
  month   = may,
  number  = {3},
  pages   = {1381--1396},
  volume  = {18},
  doi     = {10.1109/TSC.2025.3568251},
}

@InProceedings{loss2025sistema,
  author    = {Loss, Stefano and Costa, Karine and Hedigliranes, Alison and Varela, Pedro and Limão, João Pedro and Batista, Thais and Neto, Aluizio Rocha and Sabino, Daniel and Cacho, Nélio and Lopes, Frederico and Cavalcante, Everton},
  booktitle = {Anais do LII Seminário Integrado de Software e Hardware (SEMISH 2025)},
  title     = {Um sistema móvel inteligente com computação na borda para atendimento de ocorrências de segurança pública},
  year      = {2025},
  address   = {Brazil},
  note      = {in Portuguese},
  pages     = {251--262},
  publisher = {SBC},
  doi       = {10.5753/semish.2025.8405},
}

\end{document}